\documentclass[10pt,twocolumn,letterpaper]{article}

\usepackage[pagenumbers]{cvpr}      

\usepackage{amsmath,amsfonts,bm}

\def\eqref#1{equation~\ref{#1}}

\def\1{\bm{1}}

\def\va{{\bm{a}}}
\def\vb{{\bm{b}}}

\def\mI{{\bm{I}}}

\DeclareMathAlphabet{\mathsfit}{\encodingdefault}{\sfdefault}{m}{sl}
\SetMathAlphabet{\mathsfit}{bold}{\encodingdefault}{\sfdefault}{bx}{n}

\def\sT{{\mathbb{T}}}

\def\sX{{\mathbb{X}}}
\def\sY{{\mathbb{Y}}}

\newcommand{\R}{\mathbb{R}}

\DeclareMathOperator*{\argmax}{arg\,max}
\DeclareMathOperator*{\argmin}{arg\,min}

\usepackage{graphicx}
\usepackage{amsmath}
\usepackage{amssymb}
\usepackage{booktabs}
\usepackage{comment}
\usepackage{algorithm}
\usepackage{algpseudocode}

\usepackage[pagebackref,breaklinks,colorlinks]{hyperref}

\usepackage[capitalize]{cleveref}
\crefname{section}{Sec.}{Secs.}
\Crefname{section}{Section}{Sections}
\Crefname{table}{Table}{Tables}
\crefname{table}{Tab.}{Tabs.}
\usepackage{dsfont}

\def\cvprPaperID{11876} 
\def\confName{CVPR}
\def\confYear{2023}

\def\cost{{\mathcal{C}}}
\def\loss{{\mathcal{L}}}
\def\indicator{{\mathds{1}}}

\begin{document}

\title{Re-basin via implicit Sinkhorn differentiation}

\author{Fidel A. Guerrero Pe\~{n}a\\
{\tt\small fidel-alejandro.guerrero-pena@etsmtl.ca}
\and
Heitor Rapela Medeiros\\
{\tt\small heitor.rapela-medeiros.1@ens.etsmtl.ca}
\and
Thomas Dubail\\
{\tt\small thomas.dubail.1@ens.etsmtl.ca}
\and
Masih Aminbeidokhti\\
{\tt\small masih.aminbeidokhti.1@ens.etsmtl.ca}
\and
Eric Granger\\
{\tt\small eric.granger@etsmtl.ca}
\and
Marco Pedersoli\\
{\tt\small marco.pedersoli@etsmtl.ca}
\and
LIVIA, \'Ecole de Technologie Sup\'erieure\\
Montreal, Canada\\
}
\maketitle

\begin{abstract}
    The recent emergence of new algorithms for permuting models into functionally equivalent regions of the solution space has shed some light on the complexity of error surfaces, and some promising properties like mode connectivity. However, finding the right permutation is challenging, and current optimization techniques are not differentiable, which makes it difficult to integrate into a gradient-based optimization, and often leads to sub-optimal solutions.
    In this paper, we propose a Sinkhorn re-basin network with the ability to obtain the transportation plan that better suits a given objective. Unlike the current state-of-art, our method is differentiable and, therefore, easy to adapt to any task within the deep learning domain. Furthermore, we show the advantage of our re-basin method by proposing a new cost function that allows performing incremental learning by exploiting the linear mode connectivity property. The benefit of our method is compared against similar approaches from the literature, under several conditions for both optimal transport finding and linear mode connectivity. The effectiveness of our continual learning method based on re-basin is also shown for several common benchmark datasets, providing experimental results that are competitive with state-of-art results from the literature.
\end{abstract}

\section{Introduction}
\label{sec:intro}

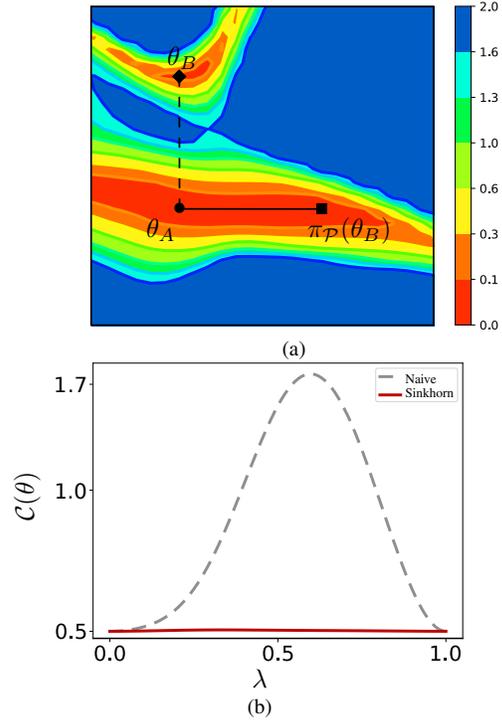
\begin{figure}[t!]
  \centering
  \begin{tabular}{c}
    \begin{subfigure}{0.65\linewidth}
      \centering
      \includegraphics[width=\linewidth]{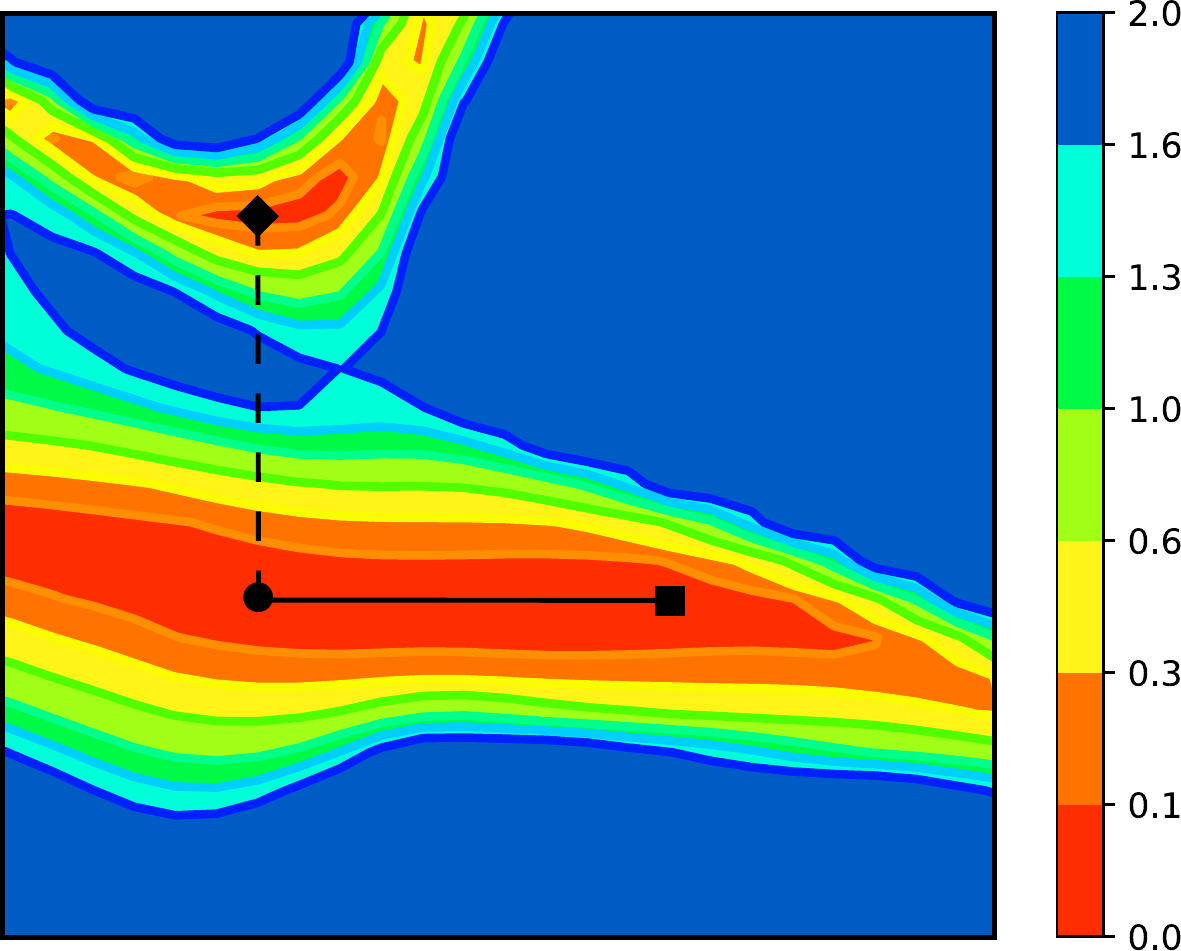}
      \put(-133,35){$\theta_A$}
      \put(-125,100){$\theta_B$}
      \put(-72,35){$\pi_\mathcal{P}(\theta_B)$}
      \caption{ }
      \label{fig:rebasin-a}
    \end{subfigure} \\
    \hspace{-10mm}
    \begin{subfigure}{0.65\linewidth}
      \centering
      \includegraphics[width=\linewidth]{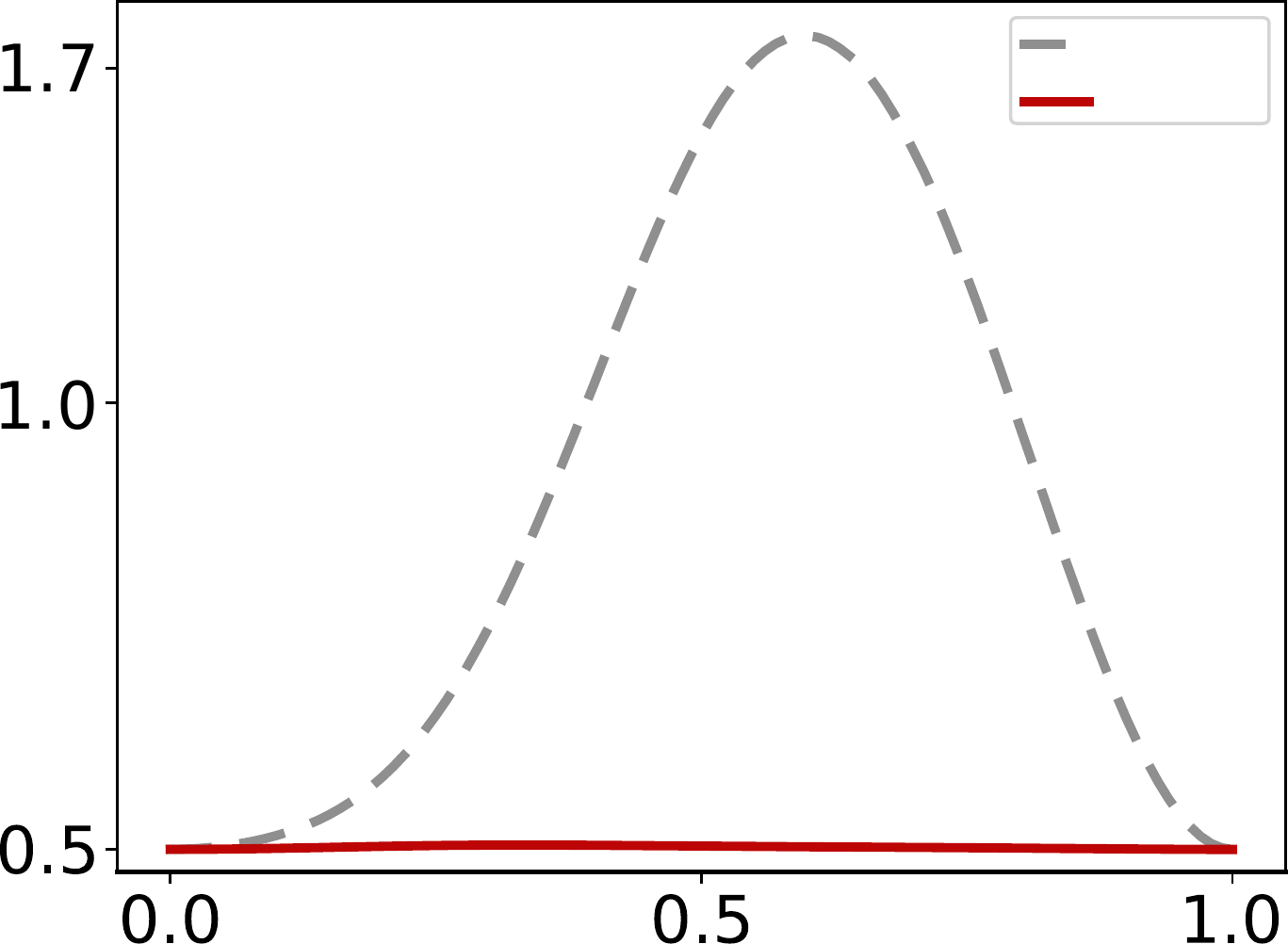}
      \put(-22,105){\tiny Naive}
      \put(-22,100){\tiny Sinkhorn}
      \put(-80,-10){$\lambda$}
      \put(-170,60){\rotatebox[origin=c]{90}{$\cost(\theta)$}}
      \caption{ }
      \label{fig:rebasin-b}
    \end{subfigure}
  \end{tabular}
  \caption{(a) The loss landscape for the polynomial approximation task \cite{oswald2020continual}. $\theta_A$ and $\theta_B$ are solutions found by SGD. LMC suggests that permuting hidden units of $\theta_B$ would result in $\pi_\mathcal{P}(\theta_B)$ which is functionally equivalent to before permutation, with no barrier on its linear interpolation with $\theta_A$. (b) Comparison of the cost value in the linear path before (naive) and after re-basin. The dashed line in both figures corresponds with the original (naive) path between models, and the solid line represents the path and corresponding loss after the proposed Sinkhorn re-basin.}
  \label{fig:rebasin}
\end{figure}

Despite the success of deep learning (DL) across many application domains, the loss surfaces of neural networks (NNs) are not well understood. Even for shallow neural networks, the number of saddle points and local optima can increase exponentially with the number of parameters \cite{auer1995exponentially, garipov2018loss}. The permutation symmetry of neurons in each layer allows the same function to be represented with many different parameter values of the same network. Symmetries imposed by these invariances help us to better understand the structure of the loss landscape \cite{garipov2018loss, entezari2022the, brea2019weight}.

Previous studies establish that minima found by Stochastic Gradient Descent (SGD) are not only connected in the network parameter's space by a path of non-increasing loss, but also permutation symmetries may allow us to linearly connect those points with no detriment to the loss \cite{entezari2022the, garipov2018loss, draxler2018essentially, kuditipudi2019explaining, frankle2020linear, nagarajan2019uniform}. This phenomenon is often referred to as linear mode connectivity (LMC) \cite{nagarajan2019uniform}. For instance, \cref{fig:rebasin-a} shows a portion of the loss landscape for the polynomial approximation task \cite{oswald2020continual} using the method proposed by Li \etal \cite{visualloss}. $\theta_A$ and $\theta_B$ are two minima found by SGD in different basins with an energy barrier between the pair. LMC suggests that if one considers permutation invariance, we can teleport solutions into a single basin where there is almost no loss barrier between different solutions \cite{entezari2022the, ainsworth2022git}. In literature, this mechanism is called re-basin \cite{ainsworth2022git}.
However, efficiently searching for permutation symmetries, that bring all solutions to one basin, is a challenging problem \cite{entezari2022the}. Three main approaches for matching units between two neural networks have been explored. The studies \cite{ainsworth2022git, singh2020model} propose a data-dependent algorithm which associates units across two NNs by matching their activations. Since activation-based matching is data dependent, it helps to adjust permutations to certain desired kinds of classes or domains \cite{singh2020model}. Instead of associating units by their activations, one could align the weights of the model itself \cite{ainsworth2022git, singh2020model}, which is independent of the dataset and therefore the computational cost is much lower. Finally, the third approach is to iteratively adjust the permutation of weights. Many studies \cite{ainsworth2022git, yurochkin2019bayesian, wang2020federated} have proposed to alternate for a number of iterations between finding an alignment and retraining to minimize the loss barriers between SGD minimas. Unfortunately, the proposed approaches so far are either non-differentiable \cite{ainsworth2022git, singh2020model, entezari2022the} or computationally expensive \cite{ainsworth2022git}, making the solution difficult to be extended to other applications, with a different objective. For instance, adapting those methods for domain adaptation by including in the optimization loss the algorithm for weight matching between two models trained on different domains is not trivial.

In this work, inspired by \cite{mena2018learning}, we relax the permutation matrix with the Sinkhorn operator \cite{adams2011ranking} and use it to solve the matching problem in a differentiable fashion. 
To avoid the high computational cost for computing gradients in Mena \etal \cite{mena2018learning} proposal, we use the implicit differentiation algorithm proposed by \cite{eisenberger2022unified} which has been shown to be more cost-effective.
Our re-basin formulation allows defining any differentiable objective as a loss function.

A direct application of re-basin is the merger of diverse models without significantly degrading their performance \cite{garipov2018loss, wang2020federated, ainsworth2022git, benzing2022random, frankle2020linear}. Applications like federate learning \cite{ainsworth2022git}, ensembling\cite{frankle2020linear}, or model initialization \cite{benzing2022random} exploit such a merger by selecting a model in the line connecting the models to be combined. Here to show the effectiveness of our approach, we propose a new continual learning algorithm that combines models trained on different domains. Our continual learning algorithm differs from previous state-of-art approaches \cite{mirzadeh2021linear} because it directly estimates a model at the intersection of previous and new knowledge by exploiting the LMC property observed in SGD-based solutions.

Our main contribution can be summarized as follows:

\noindent \textbf{(1)} Solving the re-basin for optimal transportation using implicit Sinkhorn differentiation, enabling better differentiable solutions that can be integrated on any loss. 

\noindent \textbf{(2)} A powerful way to use our re-basin method based on the Sinkhorn operator for continual learning, by considering it as a model merging problem and leveraging LMC. 

\noindent \textbf{(3)} An extensive set of experiments that validate our method for: (i) finding the optimal permutation to transform a model to another one equivalent; (ii) linear mode connectivity, to linearly connect two models such that their loss is almost identical along the entire connecting line on the weights space; and (iii) continual learning, to learn new domains and tasks while not forgetting the previous ones. 


\section{Related work}
\label{sec:related}

\noindent \textbf{Re-Basin.} Recently, in neural network community, re-basin has been demonstrating useful properties. The main goal of such re-basin approaches is to obtain functionally equivalent models in a different region of the weight space following some pre-defined objective. Permutation symmetries are a well-known example of transformations that allows performing re-basin. In particular, Entezari \etal ~\cite{entezari2022the} shows that the invariances of neural networks using random permutations on SGD solutions are likely to have almost zero barriers, and therefore the randomness in terms of permutations does not impact the quality of the final training result of the model. A simulated annealing-based algorithm was proposed for doing a re-basin with an elevated computational cost which makes it impractical to use, especially for bigger models. Ainsworth \etal~\cite{ainsworth2022git} proposed three new re-basin algorithms, that rely on solving linear assignment problems to find permutation matrices that satisfy their encoded objective. The methods shown perform well, especially in achieving linear mode connectivity. On the downside, new objectives are difficult to plug into their framework and their solution uses greedy algorithms which do not guarantee to find the optimal solution, as shown in our experiments. Using a similar approach, Benzing \etal.~\cite{benzing2022random} found strong evidence that two random initialization of a neural network after permutation can lead to a good performance, showing that the random initialization is already in the same loss valley during the initialization. Finally,~\cite{akash2022wasserstein} uses the concepts of Wasserstein Barycenter and Gromov-Wasserstein Barycenter, offering a framework for neural network model fusion with insights about linear mode connectivity of SGD solutions. Even though the previous works presented solutions to perform re-basin by solving linear assignment problems, their approach fails to generalize well for other objectives. Using gradient descent-based algorithms seems to be a more suitable approach.

\noindent \textbf{Differentiating through permutations.} Permutations are fundamental for diverse applications, especially those that involve aligning and sorting data, and are the core of most current re-basin algorithms. However, the solution space is constrained to the set of binary permutation matrices, which makes differentiation very difficult. Mena \etal \cite{mena2018learning} proposes the Gumbel-Sinkhorn, an extension of the Gumbel-Softmax method, which is a non-differentiable parametrization of a permutation that is approximated by the Sinkhorn operator. Recently, \cite{eisenberger2022unified} proposes a computationally efficient and robust approach to differentiate the Sinkhorn operator. 
We based our proposal on these studies providing solutions to obtain a differentiable re-basin via permutation. Such an approach simplifies the generalization to other applications such as continual learning and federated learning and makes it easy to integrate into neural networks.

\noindent \textbf{Mode connectivity.} Mode connectivity is responsible for demonstrating that the loss minima of different models can be connected in the weight space with almost zero loss barrier, introducing the concept of the basin of equivalent solutions. In their work, Garipov \etal \cite{garipov2018loss} found that the local optima of deep learning models are connected by simple curves. As an application for their proposal, the Fast Geometric Ensembling method was proposed. Almost at the same time, \cite{draxler2018essentially} proposed a Nudged Elastic Band-based method to construct continuous paths between minima of neural networks architectures. Finally, Frankle \etal ~\cite{frankle2020linear} studies the sensitivity of different levels of SGD noise on neural networks. These pioneering works are the basis for applications of mode connectivity, like \cite{ainsworth2022git, mirzadeh2021linear} and ours.

\noindent \textbf{Continual learning.} Continual or incremental learning (CL) has received much attention in the machine/deep learning community, allowing us to adapt models incrementally based on new training data, without forgetting previous knowledge. Catastrophic forgetting~\cite{mccloskey1989catastrophic, ratcliff1990connectionist} occurs when a model that is trained for a task on a new dataset loses information learned to perform well on that task on the original dataset. To address this issue for CL,~\cite{kirkpatrick2017overcoming} proposes Elastic Weight Consolidation (EWC), which smooths the catastrophic forgetting by regularizing neural network parameters with respect to the importance of the weights concerning the previous and actual tasks. Chaudhry \etal ~\cite{chaudhry2019tiny} proposes to use a small number of samples for replay. In their work, the Experience Replay (ER) helps the CL, even with tiny episodic memory, to  improve performance on classification tasks. Closer to our work,~\cite{mirzadeh2021linear} proposes an LMC-based method with replay. Its solution is called Mode Connectivity SGD (MC-SGD), which relies on the assumption that there is always an existing solution that incrementally solves all seen tasks, and they are connected with a linear path consisting of a low value on the loss landscape. Furthermore, MC-SGD utilizes a replay buffer to remember previous tasks for CL. Its efficiency relies on exploring the linear path of low loss to constrain learning, thus performing better than competitors, such as EWC, when less data is presented. However, such an approach has a high computational cost since it requires training independent models for the new task and merging them as a separate step with the model for previous knowledge. Also, the method has been shown to be difficult to reproduce or adapt to new benchmarks \cite{mehta2021empirical}. A compelling scenario for continual learning is the usage of a linear mode connectivity path to keep learning and adapt the model without forgetting the previous knowledge. The trade-off between the flatness of the LMC path and the direction to adapt the loss can be tuned in a way that brings more stability or plasticity depending on the target final solution.


\section{Re-basin via the Sinkhorn operator}
\label{sec:proposal}

Let $f_{\theta}(.)$ be a parameterized mapping where $\theta$ represents a vector of parameters within the solution space $\Theta\subset \R^{d}$, where $d$ is the number of parameters in $\theta$. In the deep learning context, $f$ can be seen as a neural network architecture, and $f_{\theta}$ is a model with weights $\theta$. Here, we refer to $\theta$ as a model for simplicity. Consequently, the cost (or error) of a model for a given task can be defined as $\cost(\theta)=\frac{1}{|\sT|}\sum_{(x,y)\in \sT} \loss(f_{\theta}(x),y)$, where $(x,y)\in\sT$ are input and expect output in training set $\sT$, and $\loss$ is an appropriate loss function.

A function $f$ is invariant to a transformation if and only if the obtained transformed function is functionally equivalent to the original mapping. Note that such invariances can also be found between two functions within the family of parametric functions $\{f_\theta\}_{\theta\in\Theta}$. The permutation of neurons is a well-known example of such transformation applied to neural networks that allow obtaining functionally equivalent models, i.e., $f_{\theta_A}(x)=f_{\theta_B}(x), \forall x$. These invariant models are obtained via the permutation transformation or re-basin function, here defined as $\pi\colon \Theta \to \Theta$, which shifts a model to a symmetric region of the loss landscape, $\mathcal{C}(\theta)=\mathcal{C}(\pi_{\mathcal{P}}(\theta))$. In this work $\mathcal{P}=(P_1,...,P_h)$ is a transportation plan with $P_i$ contained in the transportation polytope,
\begin{equation}
  \Pi=\{P\in \R_{+}^{m\times n}|P\indicator_m=\indicator_n,P^T\indicator_n=\indicator_m\},
\end{equation}
\noindent where $\indicator_d=(1,...,1)^{d}$. Without loss of generality, let $f_{\theta}(x)=(\ell_h\circ ... \circ \ell_1)(x)$ be a neural network defined as the composition of $h$ layers such that $\ell_i(z)=\sigma(W_iz+b_i)$. Here, the weights $W_i$ and biases $b_i$ are the parameters of the network,
$\theta= \{ W_i, b_i \}_{i=1}^h$, and $\sigma$ is a non-linear activation function. Then, the re-based model $\pi_\mathcal{P}(\theta)$ can be written as the functionally equivalent mapping:
\begin{equation}
  \ell'_i(z) = \sigma(P_iW_iP^T_{i-1}z+P_ib_i),
  \label{eq:rebasin}
\end{equation}
\noindent where $P_i\in\Pi$ is a valid permutation matrix, and $P_h=P^T_0=\mI$ is the identity matrix.

With regards to permutation invariance, Entezari \etal \cite{entezari2022the} conjectured that re-based SGD solutions are likely to have a low barrier within their linear interpolation $B(\theta_A,\theta_B)\approx 0$, where $B(.)$ is defined as:
\begin{equation*}
  B(\theta_A,\theta_B)=\sup_{\lambda} [[\cost((1-\lambda)\theta_A+\lambda\theta_B)]-
\end{equation*}
\begin{equation}
  [(1-\lambda)\cost(\theta_A)+\lambda\cost(\theta_B)]],
  \label{eq:barrier}
\end{equation}
\noindent and $\lambda\in(0,1)$. This phenomenon is known as LMC \cite{frankle2020linear}, and it is a particular case of the widely studied mode connectivity \cite{garipov2018loss,draxler2018essentially}. Notably, Ainsworth \etal \cite{ainsworth2022git} proposed three approaches that ratify the conjecture in  \cite{entezari2022the} by finding a re-based model $\pi_{\mathcal{P}}(\theta_B)$ with LMC that is near to a target model $\theta_A$. \cref{fig:rebasin} depicts the goal of such re-basin approaches.
In this figure, we can observe two solutions for the task, $\theta_A$ and $\theta_B$, found through SGD. As shown in \cref{fig:rebasin-b}, the naive path between $\theta_A$ and $\theta_B$ has higher values of the barrier within the line $(1-\lambda)\theta_A+\lambda\theta_B, \lambda\in(0,1)$. On the other hand, and consistently with the results in \cite{entezari2022the,ainsworth2022git}, our re-based model $\pi_\mathcal{P}(\theta_B)$ achieves LMC by successfully finding a transportation plan $\mathcal{P}$ that shift model $\theta_B$ to the same basin of model $\theta_A$.

Although the seminal work by Ainsworth \etal \cite{ainsworth2022git} proposed a highly efficient approach for finding a permutation that minimizes the distances between models, $\argmin_{\mathcal{P}} ||\theta_A - \pi_{\mathcal{P}}(\theta_B)||^2$, their non-differentiable approach provides solutions that are difficult to be extended to other applications with a different objective. Specifically, their algorithms use a formulation based on the linear assignment problem (LAP) to find suitable permutations, meaning any new objective needs to be cast as a LAP which is a hard task in itself.

In this work, a differentiable approach is proposed to perform re-basin that allows defining any differentiable objective as a loss function. Here, we relax the rigid constraint of having a binary permutation matrix $P$, and consequently add an entropy regularizer $h(P)=-\sum P (\log P)$ to the original LAP as proposed by \cite{mena2018learning}. The final equation is then defined as:
\begin{equation}
  S_\tau(X) = \argmax_{P\in\Pi} \langle P,X\rangle_{F}+\tau h(P),
  \label{eq:sinkhorn}
\end{equation}
being $\tau$ a factor that weights the strengths of the entropy regularization term.

The formulation in \cref{eq:sinkhorn} is known as the Sinkhorn operator and can be efficiently approximated by:
\begin{equation*}
  S_{\tau}^{(0)}(X) = \exp\left(\frac{X}{\tau}\right),
\end{equation*}
\begin{equation}
  S_{\tau}^{(t+1)}(X)= \mathcal{T}_c(\mathcal{T}_r(S_{\tau}^{(t)}(X))).
  \label{eq:sinkhorn2}
\end{equation}
where $X\in\R^{m\times n}$ is a soft version of the permutation matrix, $\mathcal{T}_c(X)=X\oslash(\indicator_{m}\indicator_{m}^{T}X)$ and $\mathcal{T}_r(X)=X\oslash(X\indicator_{n}\indicator_{n}^{T})$ are respectively the re-normalization of columns and rows of $X$, and $\oslash$ is the element-wise division. In their work, Mena \etal \cite{mena2018learning} proved that \cref{eq:sinkhorn2} converges to \cref{eq:sinkhorn} when $t\to\infty$. However, in practice, only a finite number of iterations are needed to produce a suitable approximation. 

Although the Sinkhorn operator is reasonably easy to implement within the neural network layers, a significant drawback arises when considering the efficiency of its differentiation. We use the implicit differentiation algorithm proposed by Eisenberger \etal \cite{eisenberger2022unified} to mitigate such an increase in the computational cost. Their method significantly increases the efficiency, and also stability of the training process. The marginals of the generic formulation in \cite{eisenberger2022unified} are defined as $\va=\indicator_m/m$ and $\vb=\indicator_n/n$ to match the re-basin task.

Finally, our proposed Sinkhorn re-basin re-writes the original re-based mapping in \cref{eq:rebasin} as:
\begin{equation}
  \ell'_i(z) = \sigma(S_{\tau}(P_i)W_iS_{\tau}(P^T_{i-1})z+S_{\tau}(P_i)b_i).
  \label{eq:sinkhorn-rebasing}
\end{equation}
where $P_i\in\R^{m\times n}$ is a differentiable cost matrix. In all our experiments, $P_i$ are initialized to the identity matrix. Note that in contrast with  non-differentiable approaches \cite{eisenberger2022unified,ainsworth2022git,benzing2022random}, our method can update all permutation matrices at the same time during optimization.

To show the ability of our proposed Sinkhorn re-basin to minimize any differentiable objective, we provide three examples of cost functions that are minimized without changing the formulation in \cref{eq:sinkhorn-rebasing}. These cost functions are used in the SGD framework to compute the gradient of the weights in our Sinkhorn re-basin network and finally update the permutation matrices to minimize the objective. First, for a data-free objective like Weights Matching \cite{ainsworth2022git}, we directly minimize the squared L2 distance defined as:
\begin{equation}
  \cost_{L2}(\mathcal{P};\theta_A,\theta_B) = ||\theta_A-\pi_{\mathcal{P}}(\theta_B)||^2.
  \label{eq:rebasinl2}
\end{equation}

Given the ability of our method to use differentiable objectives, we introduce two other data-driven cost functions. Inspired by the Straight-Through Estimator in \cite{ainsworth2022git}, we propose a differentiable midpoint cost function to minimize the barrier,
\begin{equation}
  \cost_{Mid}(\mathcal{P};\theta_A,\theta_B) = \cost\left(\frac{\theta_A + \pi_{\mathcal{P}}(\theta_B)}{2}\right).
  \label{eq:mid}
\end{equation}
Since minimizing the midpoint can lead to a multimodal cost path, i.e., lower cost value for $\lambda=0.5$, and higher cost values elsewhere, we propose a cost function that minimizes the cost at random points within the line:
\begin{equation}
  \cost_{Rnd}(\mathcal{P};\theta_A,\theta_B) =\cost\left((1-\lambda)\theta_A + \lambda\pi_{\mathcal{P}}(\theta_B)\right),
  \label{eq:rand}
\end{equation}
with $\lambda$ uniformly sampled at each iteration, $\lambda\sim U(0,1)$.

\section{Re-basin incremental learning}
\label{sec:proposal2}

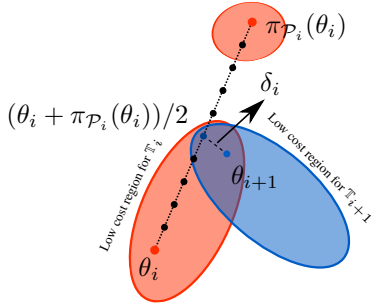
\begin{figure}[t]
  \centering
  \includegraphics[height=40mm]{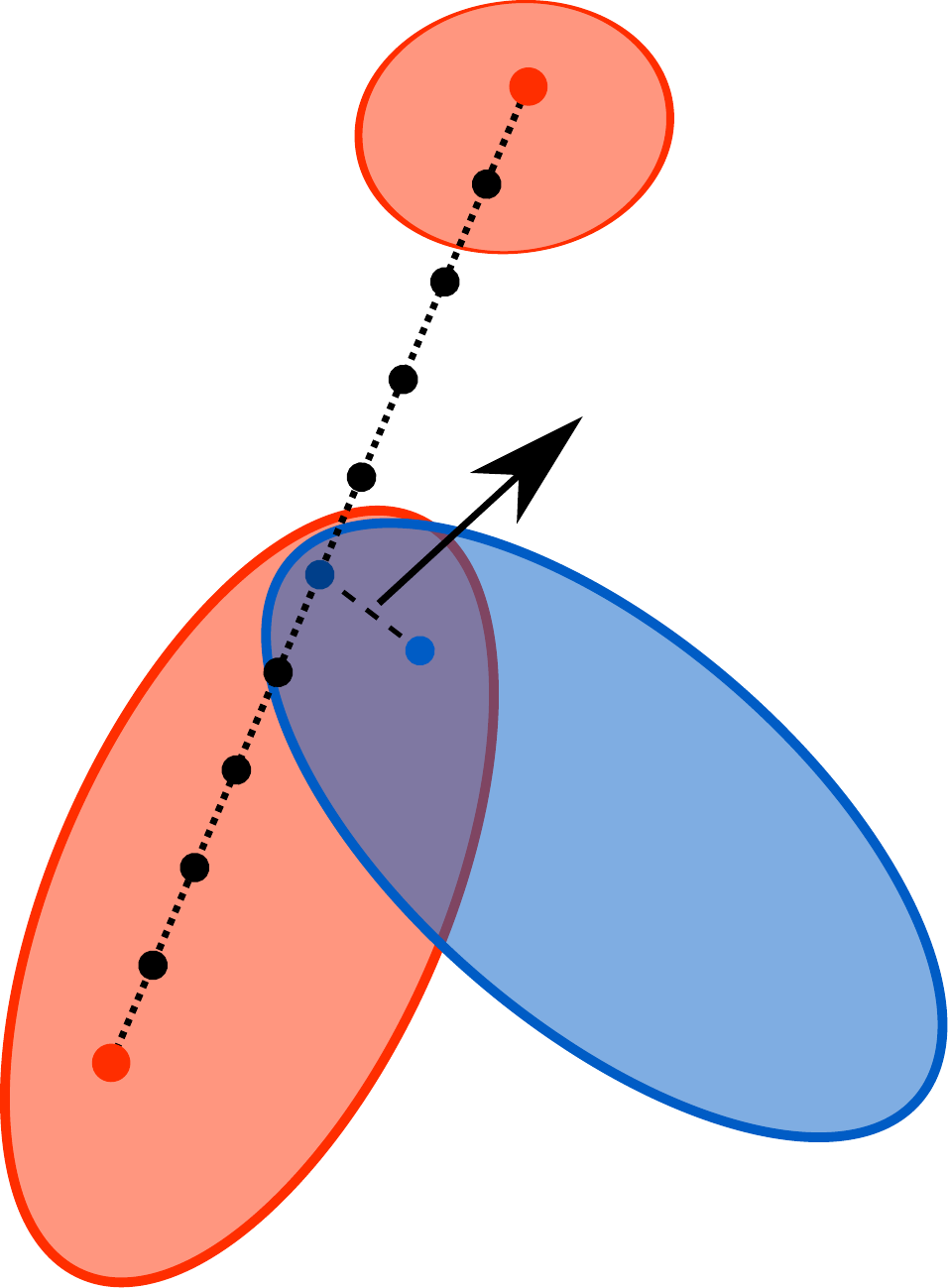}
  \put(-80,10){$\theta_i$}
  \put(-32,102){$\pi_{\mathcal{P}_i}(\theta_i)$}
  \put(-130,68){$(\theta_{i}+\pi_{\mathcal{P}_i}(\theta_i))/2$}
  \put(-45,45){$\theta_{i+1}$}
  \put(-34,80){$\delta_{i}$}
  \put(-95,40){\rotatebox[origin=c]{65}{\tiny Low cost region for $\sT_i$}}
  \put(-33,50){\rotatebox[origin=c]{-45}{\tiny Low cost region for $\sT_{i+1}$}}

  \caption{Graphical representation of the intersection of low curvature regions of the loss landscape for task $\sT_i$ (red) and $\sT_{i+1}$ (blue). The goal of our method is to find a re-basin $\pi_{\mathcal{P}_i}(\theta_i)$ that transverses the multitask region using LMC. The final model $\theta_{i+1}$ (\cref{eq:clrebasin_newmodel}) is found in the surroundings of the line by adding a learnable residual $\delta_i$.}
  \label{fig:rebasincl}
\end{figure}

A critical application of re-basin approaches is the ability to merge models without a significant performance reduction. Such a merging is usually done by selecting a model in their connecting line. Applications like federate learning \cite{ainsworth2022git}, ensembling\cite{frankle2020linear}, or model initialization \cite{benzing2022random} have been explored recently for merging models trained on the same task. Here we push the merging based on re-basin further by proposing a new incremental learning approach that fuses models trained on different domains or classes. Our proposed approach relies on a stability-plasticity hyper-parameter that allows us choosing the balance between forgetting and incorporating the new knowledge.

Let $\theta_0$ be an initial model trained over dataset $\sT_0$. Let also $\mathcal{T}=\{\sT_1, \sT_2, ...\}$ be a stream of data where $\sT_i=\{(x_{ij},y_{ij})|x_{ij}\in\sX_i, y_{ij}\in\sY_i\}, 1\leq i\leq N_i$, is a supervised dataset with input $\sX_i$, output $\sY_i$, and $N_i$ data points. Note that the sets $\sT_i$ are also known in continual learning literature as tasks but these are not limited to task incremental learning scenarios, but also include domain and class incremental learning. An incremental or continual learning process seeks to incorporate the new knowledge $\sT_{i+1}$ into the model $\theta_i$ without forgetting how to perform correctly in previous datasets $\sT_0,...,\sT_i$. To this end, 
the continual learning community has proposed approaches that exploit the fact that multitask low curvature regions usually appear at the intersection of low curvature regions for individual tasks \cite{kirkpatrick2017overcoming,mirzadeh2021linear} (see \cref{fig:rebasincl} for a visual reference). In particular, Mirzadeh \etal \cite{mirzadeh2021linear} approach uses a two steps training where the model $\theta_{i+1}$ is first trained over dataset $\sT_{i+1}$ and then a mode connectivity-based merging finds the model with low loss value on both new and previous knowledge.

In our approach, different from \cite{mirzadeh2021linear}, we directly estimate a model in the intersection of previous and new knowledge by exploiting our differentiable method to obtain the LMC observed in SGD-based solutions. Similarly to the approaches introduced in the last section, our method looks for a re-basin of the given model that minimizes a given objective. For continual learning purposes, a new cost function is introduced such that, similarly to \cref{eq:mid}, the cost of the model in the middle of the line $(1-\lambda)\theta_i+\lambda\pi_{\mathcal{P}_i}(\theta_i),\lambda\in(0,1),$ is minimized for dataset $\sT_{i+1}$, (see \cref{fig:rebasincl}). In such a continual learning scenario, the middle point is the furthest model in the line from high stability points $\theta_i$ and $\pi_{\mathcal{P}_i}(\theta_i)$. 
Constraining the solution space to the models within the line yields a solution that performs well on the previous task, but does not allow optimal performance on the new task, thus affecting the training plasticity. Similarly to Kirkpatrick \etal \cite{kirkpatrick2017overcoming}, we find well-behaved models for $\sT_{i+1}$ in the neighborhood of our optimization target. This is done by adding a residual vector $\delta_i$ with $l_2$ norm close to zero with a regularization term. Finally, the proposed cost is calculated as:
\begin{equation}
  \cost_{CL}(\delta_i, \mathcal{P}_i;\theta_i) =  \cost\left(\frac{\theta_i+\pi_{\mathcal{P}_i}(\theta_i)}{2}+\delta_i\right)+\beta||\delta_i||^2.
  \label{eq:clcost}
\end{equation}

During the learning phase, the underlying optimization problem finds:
\begin{equation}
  \delta_i^*, \mathcal{P}_i^* = \argmin_{\delta_i, \mathcal{P}_i} \cost_{CL}(\delta_i, \mathcal{P}_i;\theta_i),
\end{equation}
where $\delta_i^*$ and $\mathcal{P}_i^*$ are found at the same time. Note that the cost in \cref{eq:clcost} can be computed over any knowledge dataset. In this work, a replay method is used by taking the average of \cref{eq:clcost} for current and previous datasets. 

A fused model with a balance between previous and new knowledge should be obtained after minimizing the cost in \cref{eq:clcost}. Here, the incremented model at episode $i+1$ is defined as:
\begin{equation}
  \theta_{i+1} =  (1-\alpha)\theta_i + \alpha\pi_{\mathcal{P}_i}(\theta_i)+\delta_i,
  \label{eq:clrebasin_newmodel}
\end{equation}
where $\alpha$ is a hyper-parameter controlling the balance between plasticity and stability. Note that values of $\alpha$ near $0.5$ favors higher plasticity, while values around $0.0$ and $1.0$ give more importance to previous knowledge.

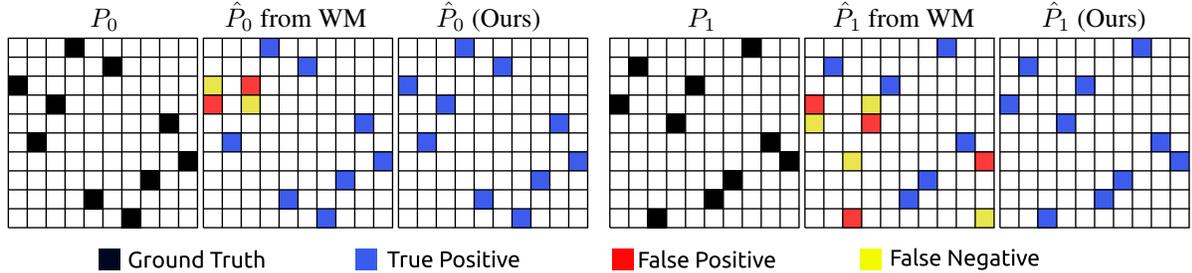
\begin{figure*}[t!]
    \centering
    \includegraphics[width=.9\linewidth]{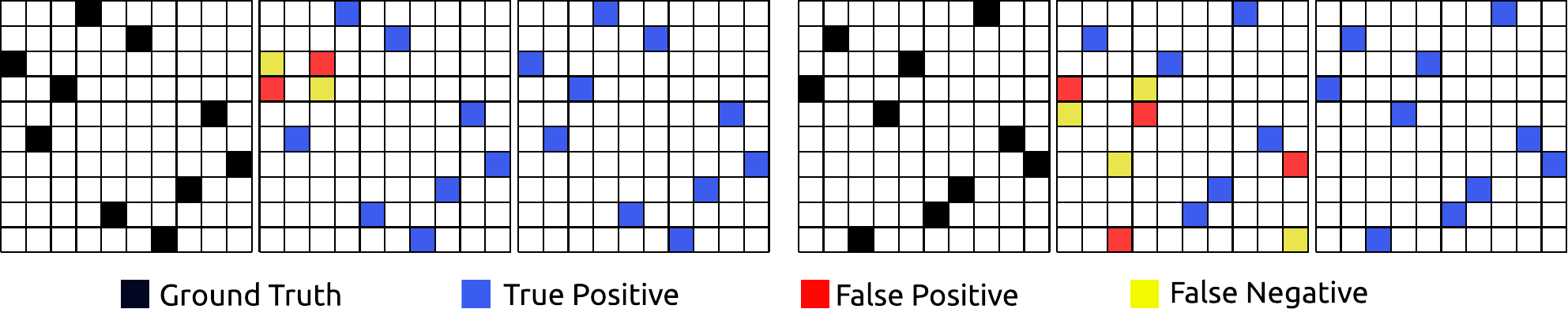}
    \put(-416,92){$P_0$}
    \put(-365,92){$\hat{P}_0$ from WM}
    \put(-285,92){$\hat{P}_0$ (Ours)}
    \put(-190,92){$P_1$}
    \put(-135,92){$\hat{P}_1$ from WM}
    \put(-56, 92){$\hat{P}_1$ (Ours)}
    \caption{Estimated permutation matrices via  Weight Matching (WM) \cite{ainsworth2022git} and the proposed Sinkhorn re-basin. $P_i$ refers to the expected $10\times 10$ permutation matrix with ones represented in black and zeros in white. The estimated permutations matrix $\hat{P}_i$ shows matching permutations as blue squares and miss-matchings in red and yellow. The permutation matrices $P_i\in\R^{10\times 10}$ correspond with transportation plans of layer $i$, with each layer containing 10 neurons. These matrices correspond with actual permutation matrices from the experiment with random initialization and 2 hidden layers.}
    \label{fig:estimated_permutations}
\end{figure*}

\section{Experimental results and analysis}
\label{sec:results}

The experimental procedure for comparing re-basin approaches follows the same one used by Ainsworth \etal \cite{ainsworth2022git}, while the continual learning experiments follow the standard experimental procedure in incremental learning literature \cite{mirzadeh2021linear}. For all experiments, the mean and standard deviation of results are reported, and were obtained over independent runs with different seeds. We used the original implementation provided by the authors in all cases. We study the effect of re-basin for both classification and regression tasks. Mnist and Cifar10 datasets were used for image classification. The polynomial approximation problems from \cite{oswald2020continual} were used for regression. As backbone architectures, we explored feedforward neural networks with 2 to 8 layers. In all our experiments, we use $t=20$ and $\tau=1.0$ as proposed in \cite{mena2018learning} for the Sinkhorn operator. Furthermore, the corresponding performance measures and hyper-parameters are summarized in each subsection. Additional details on the experimental methodology, e.g., dataset, protocol, and performance measures, are provided in the supplementary material.

\subsection{Finding the optimal transport}

In this experiment, we measure the ability of both Weights Matching (WM) \cite{ainsworth2022git} and our Sinkhorn re-basin to find the optimal permutation. Similar to our $\cost_{L2}$, the Weights Matching method minimizes the norm between the re-based model and a target model by solving a LAP. The cost function used for our re-basin is the squared L2 distance between models (\cref{eq:rebasinl2}). Note that the objectives are not data-driven, and therefore we only measure the ability of each algorithm to reach the global minima without any context. Each method received a model and a randomly permuted version of it, with the goal of finding the permutation matrices that originated the target re-basin.

For our purposes, 9 datasets are created, each one containing 50 models and a random re-basin, $\sT_i=\{(\theta_{ij},\pi_{\mathcal{P}}(\theta_{ij}))\mid P_k\sim\text{U}(\Pi), \forall P_{k}\in\mathcal{P}\}$, where $1\leq j\leq 50$ is the index of the model within dataset $\sT_i$ and $1\leq k \leq h$ is the number of hidden layers in the neural network. Note that we select random permutation matrices following a uniform distribution, $\text{U}(\Pi)$. Neural networks with two, four, and eight hidden layers were used as base architecture. Additionally, we tested three types of initializations -- random initialization with weights following a normal distribution $\mathcal{N}(0,1)$, hereafter called \emph{Rnd}, and models trained in a third and first-degree polynomial approximation problem, named \emph{Pol3} and \emph{Pol1} respectively. The 9 data set configurations lie within the combination \{Rnd, Pol3, Pol1\}$\times$\{2 hidden, 4 hidden, 8 hidden\}.

\begin{table}[t]
    \centering
    \begin{tabular}{@{}l@{\kern1.0em}l@{\kern1.0em}c@{\kern1.0em}c@{\kern1.0em}c@{}}
        \toprule
        Method                     & Init & 2 hidden $\downarrow$  & 4 hidden $\downarrow$  & 8 hidden  $\downarrow$ \\
        \midrule
        WM \cite{ainsworth2022git} & Rnd  & 6.05$\pm$9.17          & 4.12$\pm$6.58          & 0.50$\pm$1.55          \\
        $\cost_{L2}$ (Ours)        &      & \textbf{0.00$\pm$0.00} & \textbf{0.00$\pm$0.00} & \textbf{0.00$\pm$0.00} \\
        \midrule
        WM \cite{ainsworth2022git} & Pol3 & 0.57$\pm$2.84          & 0.07$\pm$0.46          & 0.01$\pm$0.10          \\
        $\cost_{L2}$ (Ours)        &      & \textbf{0.00$\pm$0.00} & \textbf{0.00$\pm$0.00} & \textbf{0.00$\pm$0.00} \\
        \midrule
        WM \cite{ainsworth2022git} & Pol1 & 0.27$\pm$0.94          & \textbf{0.00$\pm$0.00} & \textbf{0.00$\pm$0.00} \\
        $\cost_{L2}$ (Ours)        &      & \textbf{0.00$\pm$0.00} & \textbf{0.00$\pm$0.00} & \textbf{0.00$\pm$0.00} \\
        \bottomrule
    \end{tabular}
    \caption{L1 distance between the estimated and expected re-basing with different network initialization and depth. Distances are scaled $\times 10^3$. }
    \label{tab:expot}
\end{table}

\begin{table*}[t]
    \centering
    \begin{tabular}{@{}lllllllll@{}}
        \toprule
                                      & \multicolumn{2}{c}{First degree polynomial} & \multicolumn{2}{c}{Third degree polynomial} & \multicolumn{2}{c}{Mnist}            & \multicolumn{2}{c}{Cifar10}                                                                                                                                                                                  \\
        \multicolumn{1}{@{}l}{Method} & \multicolumn{1}{c}{AUC $\downarrow$}        & \multicolumn{1}{c}{Barrier $\downarrow$}    & \multicolumn{1}{c}{AUC $\downarrow$} & \multicolumn{1}{c}{Barrier $\downarrow$} & \multicolumn{1}{c}{AUC $\downarrow$} & \multicolumn{1}{c}{Barrier $\downarrow$} & \multicolumn{1}{c}{AUC $\downarrow$} & \multicolumn{1}{c}{Barrier $\downarrow$} \\
        \midrule
        Naive                         & 0.31$\pm$0.38                               & 0.62$\pm$0.71                               & 0.25$\pm$0.15                        & 0.58$\pm$0.33                            & 0.34$\pm$0.08                        & 1.07$\pm$0.21                            & 0.73$\pm$0.12                        & 1.23$\pm$0.18                            \\
        WM \cite{ainsworth2022git}    & 0.16$\pm$0.15                               & 0.32$\pm$0.28                               & 0.19$\pm$0.26                        & 0.41$\pm$0.57                            & 0.01$\pm$0.00                        & 0.03$\pm$0.01                            & 0.13$\pm$0.02                        & 0.27$\pm$0.04                            \\
        $\cost_{L2}$ (Ours)           & 0.05$\pm$0.06                               & 0.10$\pm$0.12                               & 0.05$\pm$0.06                        & 0.12$\pm$0.12                            & 0.01$\pm$0.00                        & 0.02$\pm$0.00                            & 0.09$\pm$0.02                        & 0.19$\pm$0.03                            \\
        \midrule
        STE \cite{ainsworth2022git}   & 0.11$\pm$0.10                               & 0.23$\pm$0.22                               & 0.09$\pm$0.07                        & 0.24$\pm$0.23                            & 0.01$\pm$0.00                        & 0.01$\pm$0.01                            & 0.08$\pm$0.01                        & 0.15$\pm$0.02                            \\
        $\cost_{Mid}$ (Ours)          & 0.03$\pm$0.02                               & 0.07$\pm$0.05                               & 0.05$\pm$0.04                        & 0.17$\pm$0.17                            & \textbf{0.00$\pm$0.00}               & \textbf{0.00$\pm$0.00}                   & \textbf{0.02$\pm$0.01}               & \textbf{0.05$\pm$0.01}                   \\
        $\cost_{Rnd}$ (Ours)          & \textbf{0.01$\pm$0.01}                      & \textbf{0.03$\pm$0.02}                      & \textbf{0.01$\pm$0.01}               & \textbf{0.03$\pm$0.03}                   & \textbf{0.00$\pm$0.00}               & 0.01$\pm$0.00                            & \textbf{0.02$\pm$0.01}               & 0.06$\pm$0.01                            \\
        \bottomrule
    \end{tabular}
    \caption{AUC and loss Barrier results of linear mode connectivity for regression datasets (first and third-degree polynomial approximation), and classification datasets (Mnist and Cifar10). The WM method and our Sinkhorn with $\cost_{L2}$ belong to the data-free category, while Straight-Trough Estimator STE, Sinkhorn with $\cost_{Mid}$ and $\cost_{Rnd}$ are data-driven.}
    \label{tab:exp2}
\end{table*}

The Sinkhorn re-basin model was updated using the Adam optimizer with an initial learning rate of $0.1$, and for a maximum of 5 iterations, using early stopping in case of convergence. \cref{tab:expot} summarizes the L1 norm between weights after re-basin, $|\pi_{\hat{\mathcal{P}}}(\theta)-\pi_{\mathcal{P}}(\theta)|$, where $\hat{\mathcal{P}}$ and $\mathcal{P}$ are the estimated and optimal transportation plan, respectively. As expected, our proposal always finds the optimal permutation thanks to its ability to look at all permutation matrices simultaneously. In contrast, the WM results fall short for some scenarios. It is worth mentioning that these results match the ones obtained by Ainsworth \etal in \cite{ainsworth2022git}. In general, the WM algorithm seems to be affected by random initialization, while increasing the network's capacity improves its ability to reach the global minimum. We hypothesize that this is an effect of using a greedy algorithm that optimizes the objective for different layers at each iteration. A deeper inspection of the estimated permutation matrices show that WM reaches local minima close to the expected re-basin, with only a few misplaced permutations (see \cref{fig:estimated_permutations}).

\subsection{Linear mode connectivity}

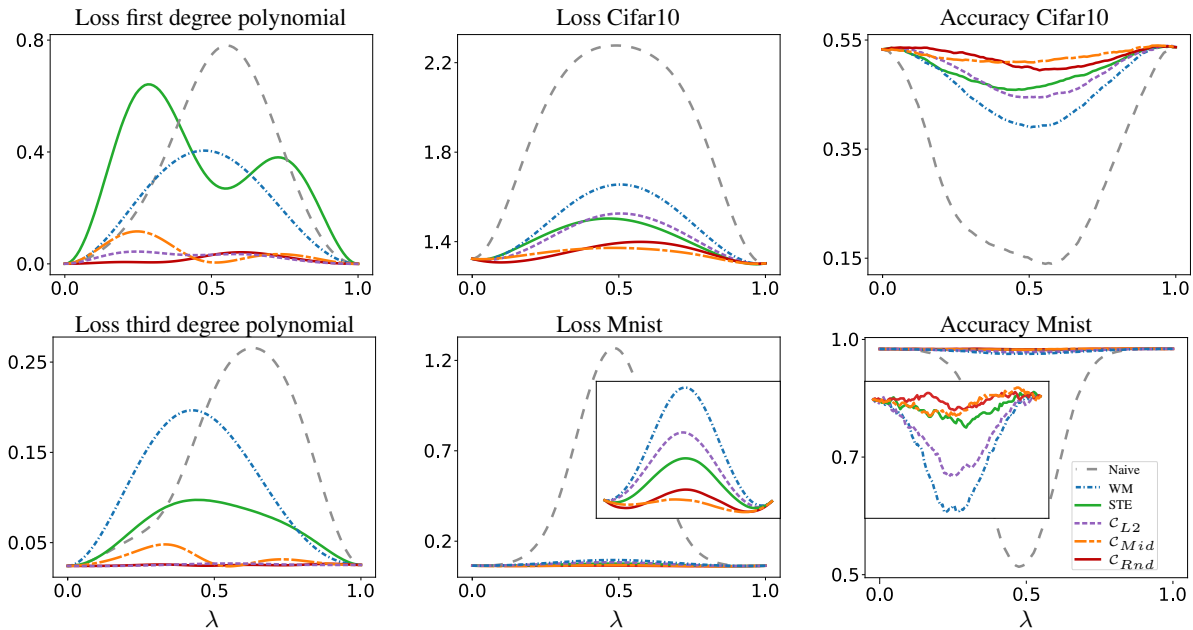
\begin{figure*}[t]

    \centering
    \includegraphics[width=0.9\linewidth]{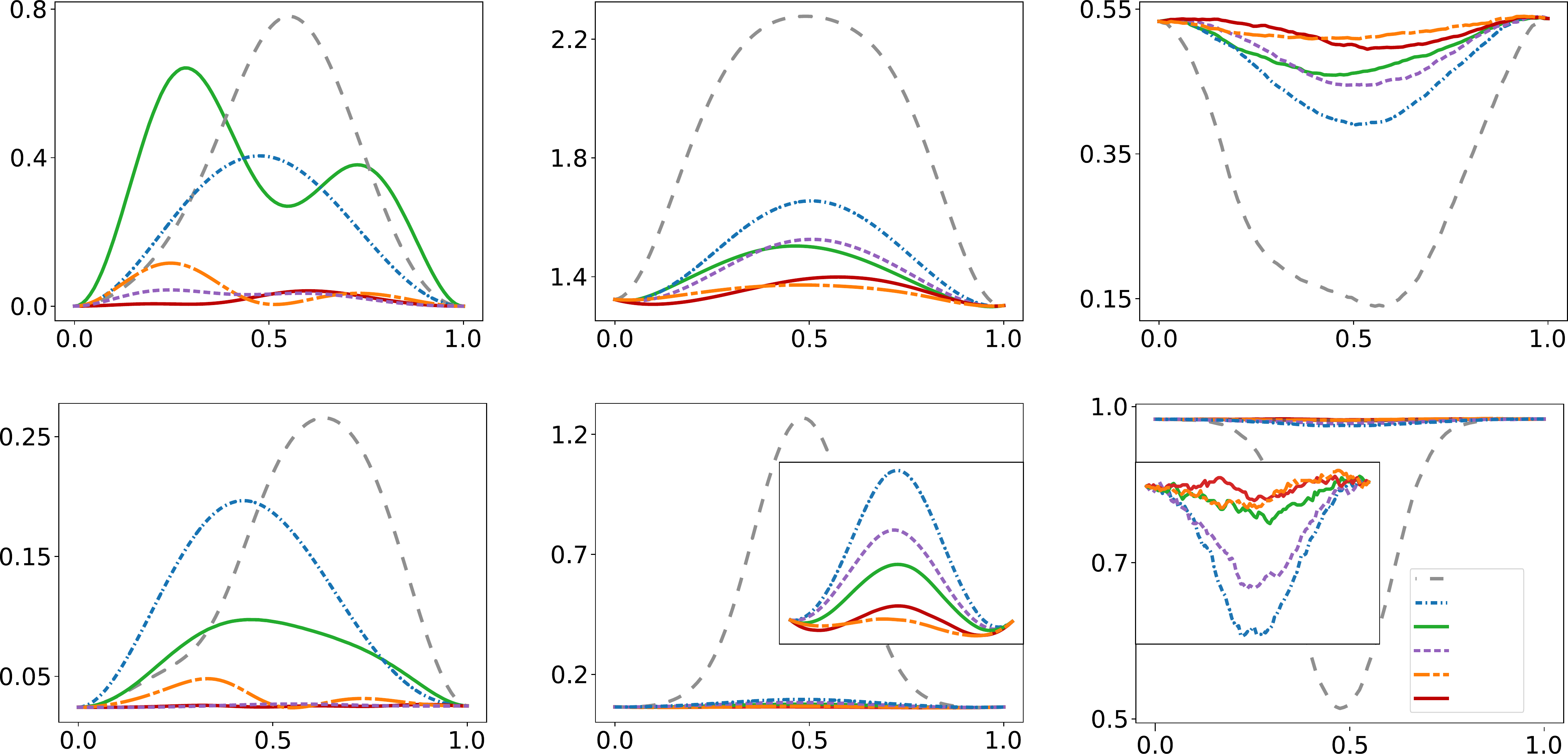}
    \put(-31,48){\tiny Naive}
    \put(-31,41){\tiny WM}
    \put(-31,35){\tiny STE}
    \put(-31,28){\tiny $\cost_{L2}$}
    \put(-31,21){\tiny $\cost_{Mid}$}
    \put(-31,14){\tiny $\cost_{Rnd}$}
    \put(-65,-10){\small $\lambda$}
    \put(-220,-10){\small $\lambda$}
    \put(-373,-10){\small $\lambda$}
    \put(-95,102){\small Accuracy Mnist}
    \put(-240,102){\small Loss Mnist}
    \put(-422,102){\small Loss third degree polynomial}
    \put(-95,218){\small Accuracy Cifar10}
    \put(-240,218){\small Loss Cifar10}
    \put(-422,218){\small Loss first degree polynomial}
    \caption{Example of linear mode connectivity achieved by Weight Matching (WM) \cite{ainsworth2022git}, Straight-Through Estimator (STE) \cite{ainsworth2022git}, and our Sinkhorn re-basin with $\cost_{L2}$, $\cost_{Mid}$, and $\cost_{Rnd}$ costs for a neural network with two hidden layers. Accuracy and loss are shown for the Mnist and Cifar10 classification, while only the L2 loss is shown for the regression tasks. For Mnist, we include an amplified version of the loss and accuracy for better comparison.}
    \label{fig:LMC}
\end{figure*}

To verify the conjecture of Entezari \etal \cite{entezari2022the}, we measure the ability of our method to find linear connectivity between SGD modes after re-basin one of them. For this experiment, four datasets are employed -- first and third-degree polynomial regression tasks \cite{oswald2020continual}, along with the classical classification benchmarks, Mnist and Cifar10. Our experiment follows a similar setup to the one described by \cite{ainsworth2022git}, i.e., two networks were trained over the same dataset, and we performed the re-basin of one of them, hoping to reach the same basin as the unchanged model. The experiment was repeated 50 times for every dataset, and each method saw the same two networks. To measure the ability of the different approaches to find LMC, we use the Barrier \cite{frankle2020linear} (\cref{eq:barrier}) and Area Under the Curve (AUC) over the estimated cost curve within the linear path.
With both measures, the best performance is achieved when a method provides a low value, with a lower bound of 0.

We compare our approach with different objectives -- L2 (\cref{eq:rebasinl2}), Middle point (\cref{eq:mid}), and Random lambda (\cref{eq:rand}) with the recently introduced Weights Matching (WM) and Straight-Through Estimator (STE) \cite{ainsworth2022git}. \cref{tab:exp2} reports the performance of the methods for each dataset. All experiments used a neural network with two hidden layers. As seen in the table, our methods outperform state-of-art methods with (first three rows) and without considering the data (last three rows) during re-basin. Specifically, our $\cost_{L2}$ method exceeded WM for both AUC and Barrier measures for all datasets, except Mnist where no significant differences were observed. In the data-driven category, our other proposals outperform the state-of-art STE approach for AUC and Barrier. In particular, our $\cost_{Rnd}$ loss showed the best results, comparable to our $\cost_{Mid}$ for more challenging scenarios like Cifar10. As a general point, all methods provide a significant improvement over the naive path.

We show the obtained loss and accuracy curves over the linear path before doing a re-basin (naive) and after applying WM, STE, and Sinkhorn with $\cost_{L2}$, $\cost_{Mid}$, and $\cost_{Rnd}$ methods in \cref{fig:LMC}. An interesting observation is that $\cost_{Mid}$ tends to find multi-modal cost curves with a valley at $\lambda=0.5$, similarly to STE.

\subsection{Incremental learning application}

\begin{table*}[h!]
    \centering
    \begin{tabular}{@{}lllllll@{}}
        \toprule
                                             & \multicolumn{2}{c}{Rotated Mnist}       & \multicolumn{2}{c}{Split Cifar100}                                                                                                  \\
        \multicolumn{1}{@{}l}{Method}        & \multicolumn{1}{c}{Accuracy $\uparrow$} & \multicolumn{1}{c}{Forgetting $\downarrow$} & \multicolumn{1}{c}{Accuracy $\uparrow$} & \multicolumn{1}{c}{Forgetting $\downarrow$} \\
        \midrule
        Finetune                             & 46.28$\pm$1.01                          & 0.52$\pm$0.01                               & 35.41$\pm$0.95                          & 0.49$\pm$0.01                               \\
        EWC~\cite{kirkpatrick2017overcoming} & 59.92$\pm$1.71                          & 0.34$\pm$0.02                               & 50.50$\pm$1.33                          & 0.24$\pm$0.02                               \\
        LwF~\cite{li2017learning}            & 61.86$\pm$3.66                          & 0.29$\pm$0.06                               & 41.43$\pm$4.06                          & 0.51$\pm$0.01                               \\
        A-GEM~\cite{chaudhry2018efficient}   & 68.47$\pm$0.90                          & 0.28$\pm$0.01                               & 44.42$\pm$1.46                          & 0.36$\pm$0.01                               \\
        Rebasin /w replay (Ours)             & \textbf{78.14$\pm$0.50}                 & \textbf{0.12$\pm$0.01}                      & \textbf{51.34$\pm$0.74}                 & \textbf{0.07$\pm$0.02}                      \\
        \midrule
        Joint training                       & 90.84$\pm$4.30                          & 0.00                                        & 60.48$\pm$0.54                          & 0.00                                        \\
        \bottomrule
    \end{tabular}
    \caption{Performance of our proposed and state-of-art methods on the continual learning benchmark datasets over 20 episodes.}
    \label{tab:cl}
\end{table*}

Although several authors have proposed methods to combine models trained over subsets of some domain, e.g., federate learning \cite{ainsworth2022git}, ensembling \cite{frankle2020linear}, and model initializations \cite{benzing2022random}, this paper explores the idea of obtaining a model that can learn a new knowledge without forgetting the previous one. To this end, this experiment seeks to compare our method with other well-known and state-of-art continual learning approaches from the literature. Since our proposal can fit the regularization techniques that use replay, we compare it with different algorithms within this category. In particular, we compared with 3 regularization-based approaches -- elastic weight consolidation (EWC) \cite{kirkpatrick2017overcoming}, learning without forgetting (LwF) \cite{li2017learning}, and average gradient episodic memory (A-GEM) \cite{chaudhry2018efficient}. The average accuracy
was calculated to measure the overall performance of model $\theta_i$ in the first $E$ episodes $\sT_{j}$. In addition, the forgetting measure
averages the forgetting in terms of accuracy for each domain or task in episode $E$.

Given the variety of libraries and implementations, we limited our comparison to reproducible models that could be used in the Avalanche environment \cite{lomonaco2021avalanche}. All measures, benchmarks, networks, and algorithms, including our own, were implemented using this framework. While we attempted to  incorporate other recent approaches like MC-SGD\cite{mirzadeh2021linear} and Stable SGD \cite{mirzadeh2020understanding} into Avalanche, a high discrepancy was observed w.r.t. their reported results and, therefore, we restrain ourselves from including them in our study. Difficulties in adapting MC-SGD to new conditions has also been observed by other authors \cite{mehta2021empirical}.

We focused our experiments on low episodic memory scenarios, using only five examples per class for both benchmarks in methods that rely on memory replay (A-GEM and our method). Similar to \cite{mirzadeh2021linear}, we used a neural network with two hidden layers and 256 neurons for the Rotated Mnist benchmark. For the Split Cifar100 benchmark, a multi-head resnet18 was used following the settings in \cite{mirzadeh2021linear,mirzadeh2020understanding}. Since our current Sinkhorn re-basin network implementation does not support residual connections, we only apply the re-basin to linear layers.

\cref{tab:cl} shows the accuracy and forgetting performance of methods on the benchmarks Rotated Mnist and Split Cifar-100 datasets using 20 episodes. Our method outperforms the others, and still achieves better or comparable performance than those reported in \cite{mirzadeh2020understanding}. The reader should pay special attention to the low values of forgetting achieved using our re-basin approach. This is a consequence of setting the value of  $\alpha=0.8$ when fusing the models (\cref{eq:clrebasin_newmodel}). A-GEM is ranked second in accuracy and forgetting for Rotated Mnist. LwF showed a similar forgetting to A-GEM in this benchmark. On Split Cifar100, EWC ranked second for both measures. Despite having similar accuracy to our approach, the high forgetting value suggests stability issues.

\section{Conclusion}
\label{sec:concl}

In this work, a new method based on the implicit Sinkhorn operator is proposed that estimates a permutation matrix which makes two neural network models equivalent. With respect to previous work, such as weight matching, our method is: (i) more flexible, because it is differentiable and can be applied with any loss, (ii) estimates the permutations for all layers at the same time, avoiding getting stuck in local minima, (iii) more accurate, as shown in our experimental evaluation on well-known benchmarks. First, our experiments yield perfect results when our approach was evaluated to produce the optimal permutation between a model and its artificially permuted transformation.  We have also used our approach for linear mode connectivity, showing better connectivity (lower loss barrier) than weight matching.  Finally, we showed that our efficient and differentiable approach for re-basin can easily be applied to the challenging task of continual learning, producing results that are comparable to, or better than state-of-art approaches. As a limitation to our work, we observed from our experiments and analysis of the literature that linear assignment problems solved with greedy Hungarian-based approaches are generally more efficient in terms of memory than the Sinkhorn operator.



{\small
    \bibliographystyle{ieee_fullname}
    \bibliography{egbib}
}

\appendix

\section{Supplementary Material}

In this appendix, we provide the details for reproducing our work. The \href{https://worksheets.codalab.org/worksheets/0x641008eb0b1b4768b865b58eddbe419c}{source code} to reproduce our results is anonymously provided. Our implementation focuses on simplicity, allowing us to pass a PyTorch module\footnote{Currently, our re-basin network only handles NNs and CNNs without skip connections. Batch normalization is supported for re-basin but not for LMC.} as an argument to our re-basin network and performing re-basin using a standard training cycle. Any cost function can be defined to guide the re-basin. The experimental setup for every experiment in the manuscript is detailed in the following sections.
\section{Finding optimal transport}

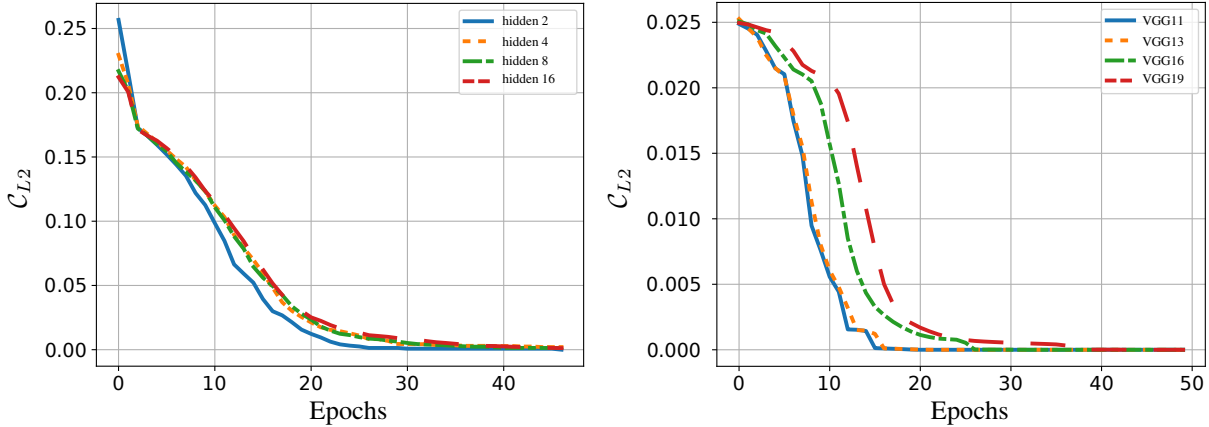
\begin{figure*}[t!]
    \centering
    \begin{tabular}{ccc}
        \includegraphics[width=.41\linewidth]{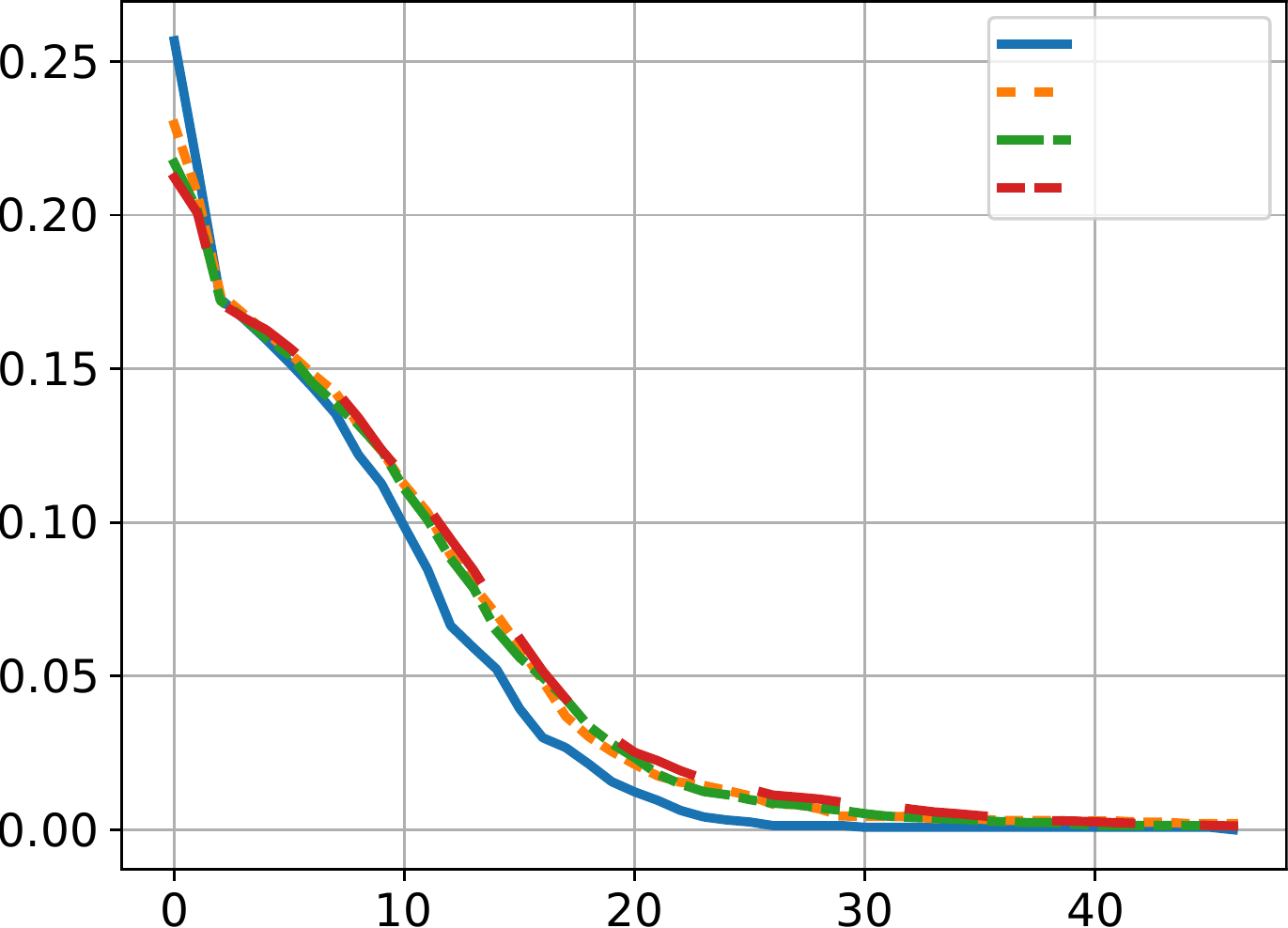} &  &
        \includegraphics[width=.425\linewidth]{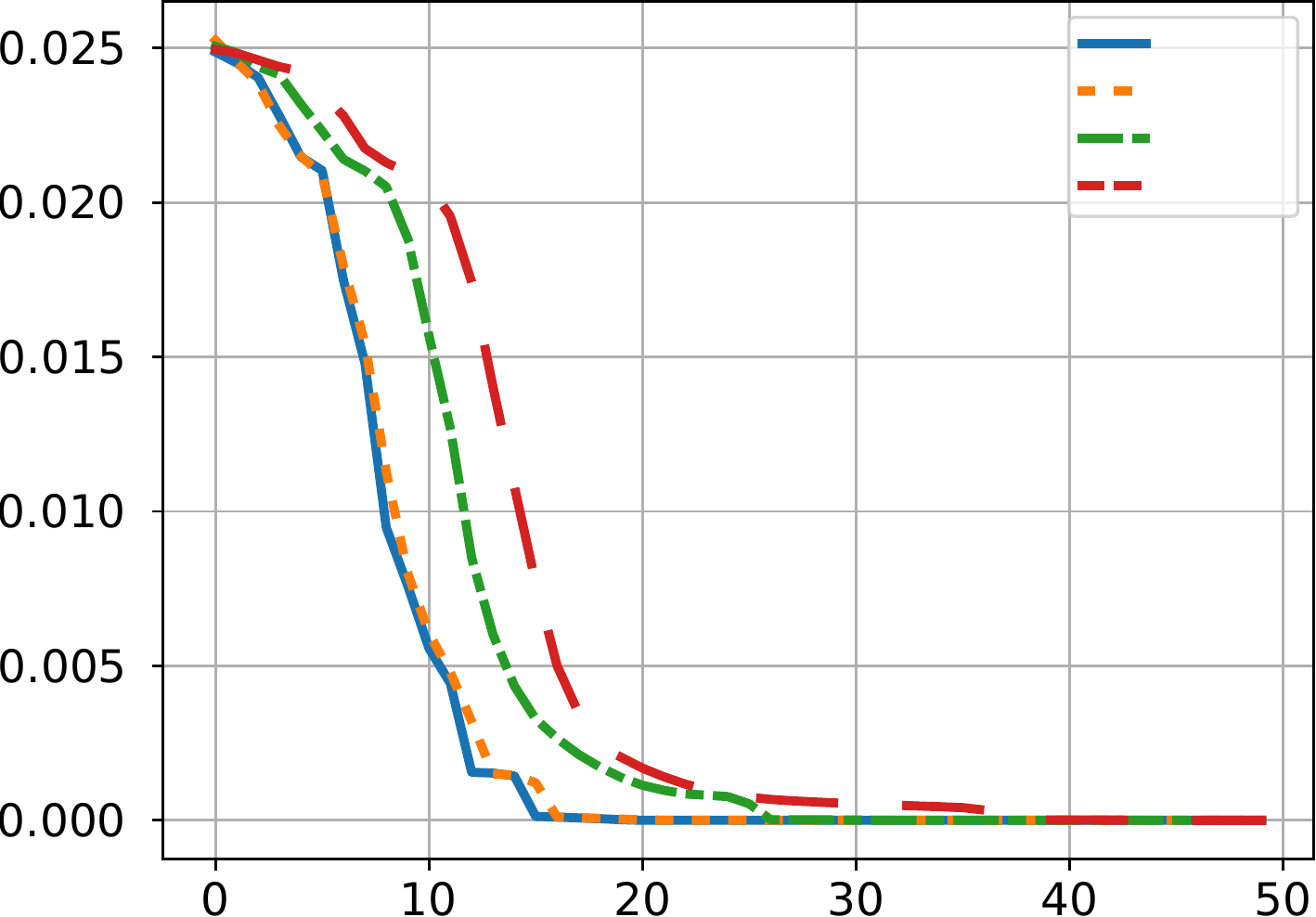}
    \end{tabular}
    \put(-345,-80){Epochs}
    \put(-459,0){\rotatebox{90}{$\cost_{L2}$}}
    \put(-110,-80){Epochs}
    \put(-232,0){\rotatebox{90}{$\cost_{L2}$}}
    \put(-272,69){\tiny hidden 2}
    \put(-272,61){\tiny hidden 4}
    \put(-272,54){\tiny hidden 8}
    \put(-272,47){\tiny hidden 16}
    \put(-30,69){\tiny VGG11}
    \put(-30,61){\tiny VGG13}
    \put(-30,54){\tiny VGG16}
    \put(-30,47){\tiny VGG19}
    \caption{Validation loss during Sinkhorn re-basin training for feedforward neural networks with a different number of hidden layers (left panel) and VGG with increasing depth (right panel).}
    \label{fig:convergence}
\end{figure*}

We used a feedforward neural network with 2, 4, and 8 hidden layers containing 10 neurons each to find the optimal transport. The hyperbolic tangent was used as an activation function. The number of inputs and outputs is set to 1 following the polynomial approximation dataset requirements. The re-basin optimization used Adam \cite{kingma2015} with an initial learning rate of 0.1. The method used the proposed data-free squared L2 distance cost function (\cref{eq:rebasinl2}). The maximum number of iterations was set to 100. However, in practice, all methods converged to 0 loss in the validation set in less than 50 iterations. An early stopping strategy was implemented to avoid running after the convergence point (see \cref{fig:convergence}). The performance measurement employed during evaluation (see Table 1 of the manuscript) was the L1 norm between the re-based model $\theta_A$ and the target model $\theta_B$:

\begin{equation}
    L1(\hat{\mathcal{P}}; \theta_A, \theta_B) = |\pi_{\hat{\mathcal{P}}}(\theta_A)-\theta_B|,
\end{equation}
with $\theta_A,\theta_B\in\R^d$, being $d$ the number of parameters in the network and $\hat{\mathcal{P}}$ the set of estimated permutation matrices.

To assess the initialization impact, we propose three initial settings for re-basin. The first one uses random initializations where the network's parameters were initialized following a normal distribution $\mathcal{N}(0,1)$. The other two settings involved training neural networks to perform regression tasks. The selected tasks were the first and third-degree polynomial approximation datasets\cite{oswald2020continual}, $\sT_{Pol1}=\{(x,y)\mid y=x+3, x\in(-4,-2)\}$ and $\sT_{Pol3}=\{(x,y)\mid y=(x-3)^3, x\in(2,4)\}$. For training the polynomial approximation networks, a small gaussian noise with distribution $\mathcal{N}(0,0.05)$ was added to the regression target \cite{oswald2020continual}.

Similar to others in the literature, our re-basin method is not limited to regression tasks or shallow feedforward neural networks. Although we selected such a scenario for comparison purposes, we want to re-enforce our approach's ability to perform well in these scenarios for classification tasks, deepest feedforward networks (NNs), and convolutional neural networks (CNNs). In particular, we use different standards of VGG without batch normalization. \cref{fig:convergence} shows the loss curves of our Sinkhorn re-basin training using the cost in \cref{eq:rebasinl2}. The initial models were obtained by training both NNs and CNNs over the Mnist dataset. The optimal transport was found in all cases in less than 20 seconds using an Nvidia GeForce RTX 3070 GPU. The source code for reproducing the experiment is provided anonymously \href{https://worksheets.codalab.org/bundles/0x5e77ad56e6c24487bbf832fac4d7ba9c}{here}. In particular, we provide examples of the execution command for both \href{https://worksheets.codalab.org/bundles/0x530c0b96887b44a99d90e82706d6400f}{NNs} and \href{https://worksheets.codalab.org/bundles/0xcf5f643bd9424cd29639e76b9dd4b328}{CNNs}.

\section{Linear mode connectivity}

To assess the capacity of achieving linear mode connectivity, we employed two classification datasets --Mnist and Cifar10--, and the previously described first and third-degree polynomial approximation datasets\cite{oswald2020continual}. As base architecture, we used a feedforward neural network with two hidden layers. In classification cases, the activation function was ReLU with $784=28\times 28$ neurons within the input layer for Mnist and $3072=32\times 32\times 3$ for Cifar10. For both benchmarks, the output layer has 10 neurons corresponding with the number of classes. The cross-entropy loss function was used for training these networks. As for the regression tasks, only 1 input and 1 output were required. Similarly to the previous experiment, hyperbolic tangent activation was used. The L2 loss function was used for training the regression networks.

The Sinkhorn re-basin networks used the Adam optimizer with a maximum of 1000 iterations. In practice, none of the executions ran the maximum number of iterations, thanks to the early stopping. In the classification settings, all experiments converged in less than 50 iterations. The best initial learning rate for every configuration is given in \cref{tab:lr2}. The mini-batch sizes were 100 for regression and 1000 for classification.

\begin{table}[h!]
    \centering
    \begin{tabular}{lccc}
        \midrule
        Dataset/Method          & $\cost_{L2}$ & $\cost_{Mid}$ & $\cost_{Rnd}$ \\
        \midrule
        First degree polynomial & 0.10         & 0.10          & 0.01          \\
        Third degree polynomial & 0.10         & 0.10          & 0.01          \\
        Mnist                   & 0.01         & 0.10          & 0.10          \\
        Cifar10                 & 0.01         & 0.10          & 0.10          \\
        \midrule
    \end{tabular}
    \caption{Initial learning rate for every dataset and method in the linear mode connectivity experiment.}
    \label{tab:lr2}
\end{table}

As performance measurement, we use the Barrier \cite{frankle2020linear}:
\begin{equation*}
    B(\theta_A,\theta_B)=\sup_{\lambda} [[\cost((1-\lambda)\theta_A+\lambda\theta_B)]-
\end{equation*}
\begin{equation}
    [(1-\lambda)\cost(\theta_A)+\lambda\cost(\theta_B)]],
\end{equation}
with $\lambda\in(0,1)$, and Area Under the Curve (AUC) over the estimated cost curve within the linear path:

\begin{equation*}
    AUC(\theta_A,\theta_B) = \int_{\lambda=0}^{1}[\cost((1-\lambda)\theta_A+\lambda\theta_B)]-
\end{equation*}
\begin{equation}
    [(1-\lambda)\cost(\theta_A)+\lambda\cost(\theta_B)]\text{d}\lambda.
\end{equation}

\begin{figure*}[t!]
    \centering
    \begin{tabular}{ccc}
        VGG11: loss                                                         &  & VGG11: accuracy \\
        \includegraphics[width=.425\linewidth]{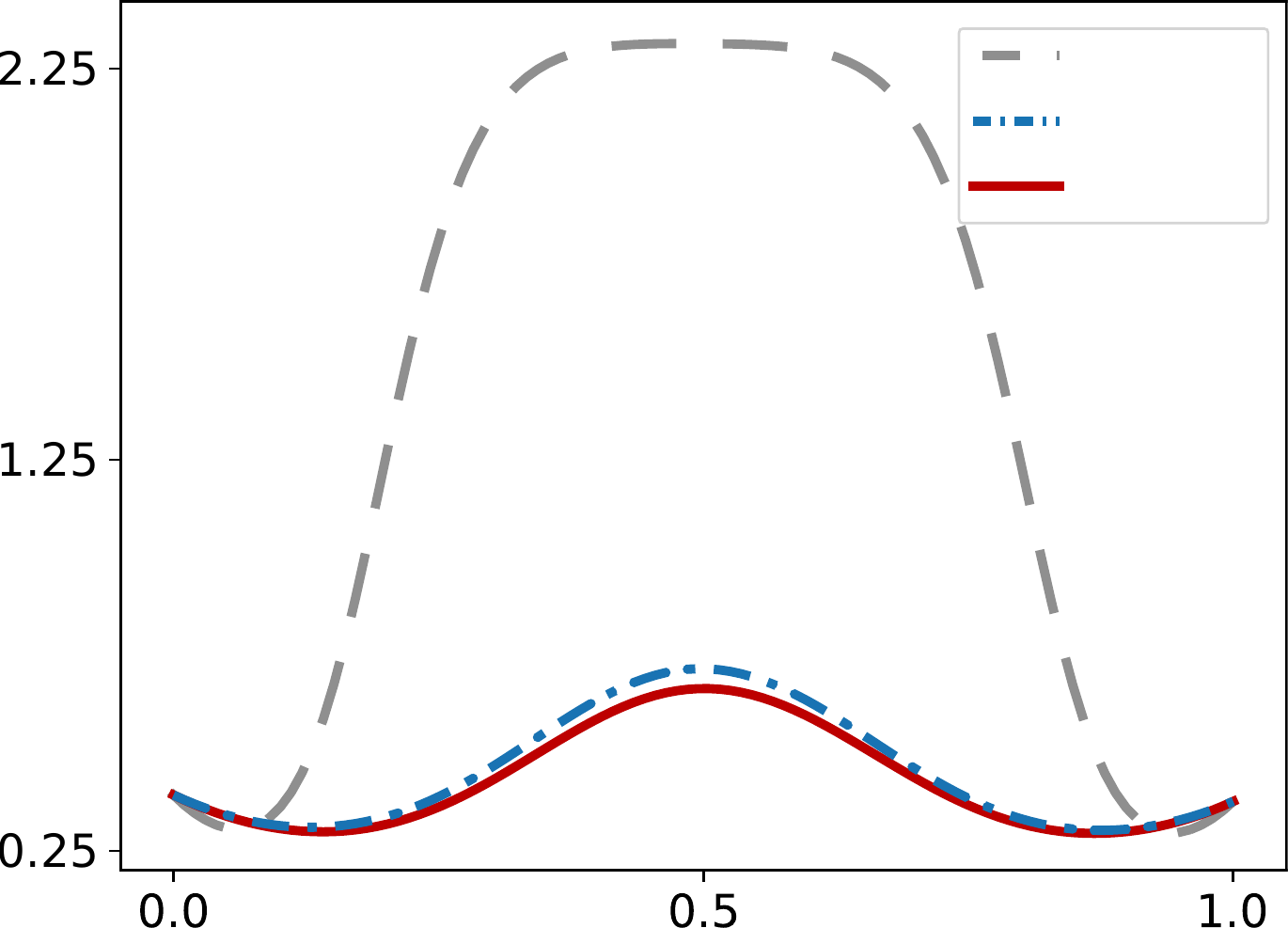} &  &
        \includegraphics[width=.41\linewidth]{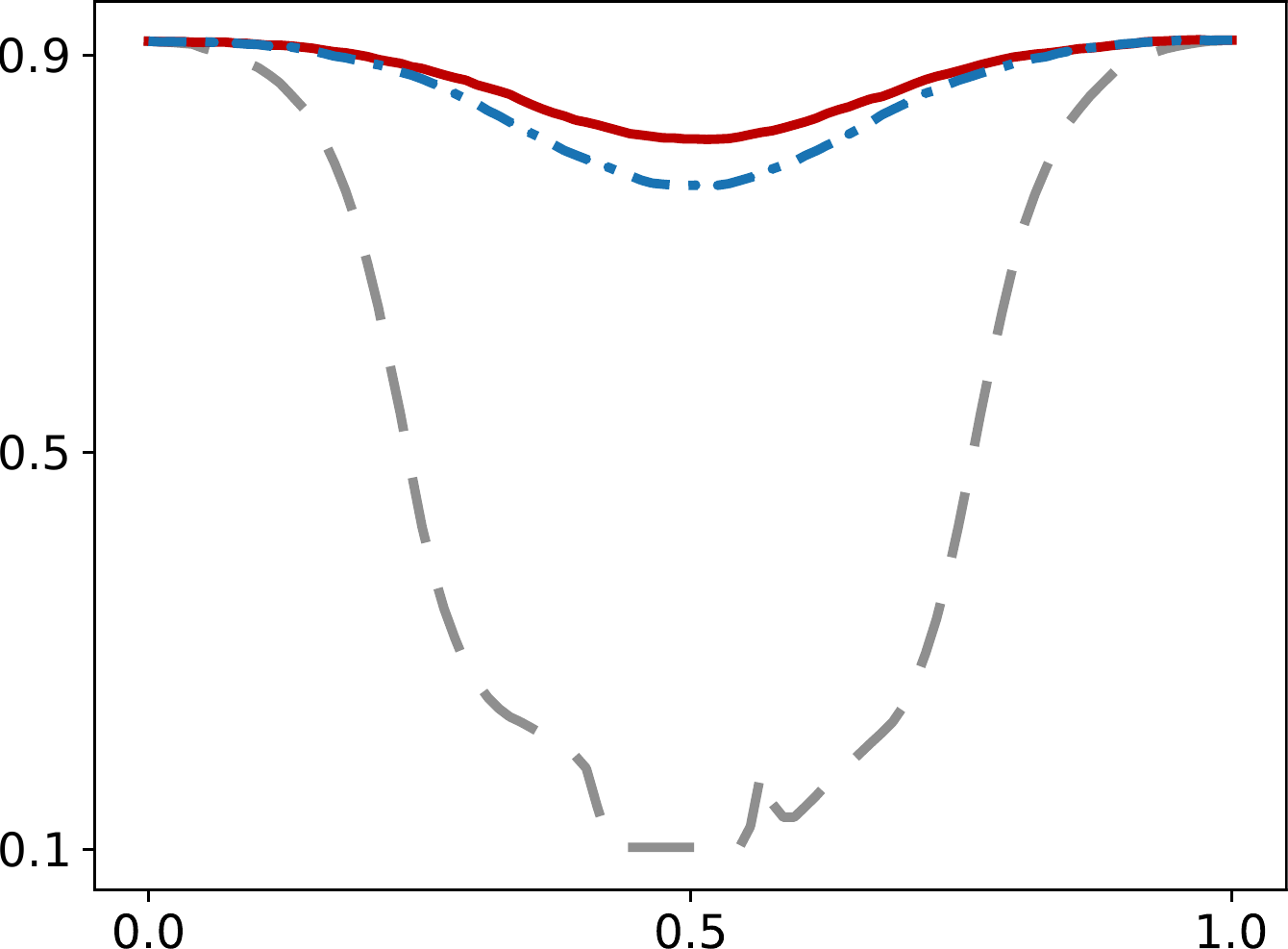}                        \\
        VGG19: loss                                                         &  & VGG19: accuracy \\
        \includegraphics[width=.425\linewidth]{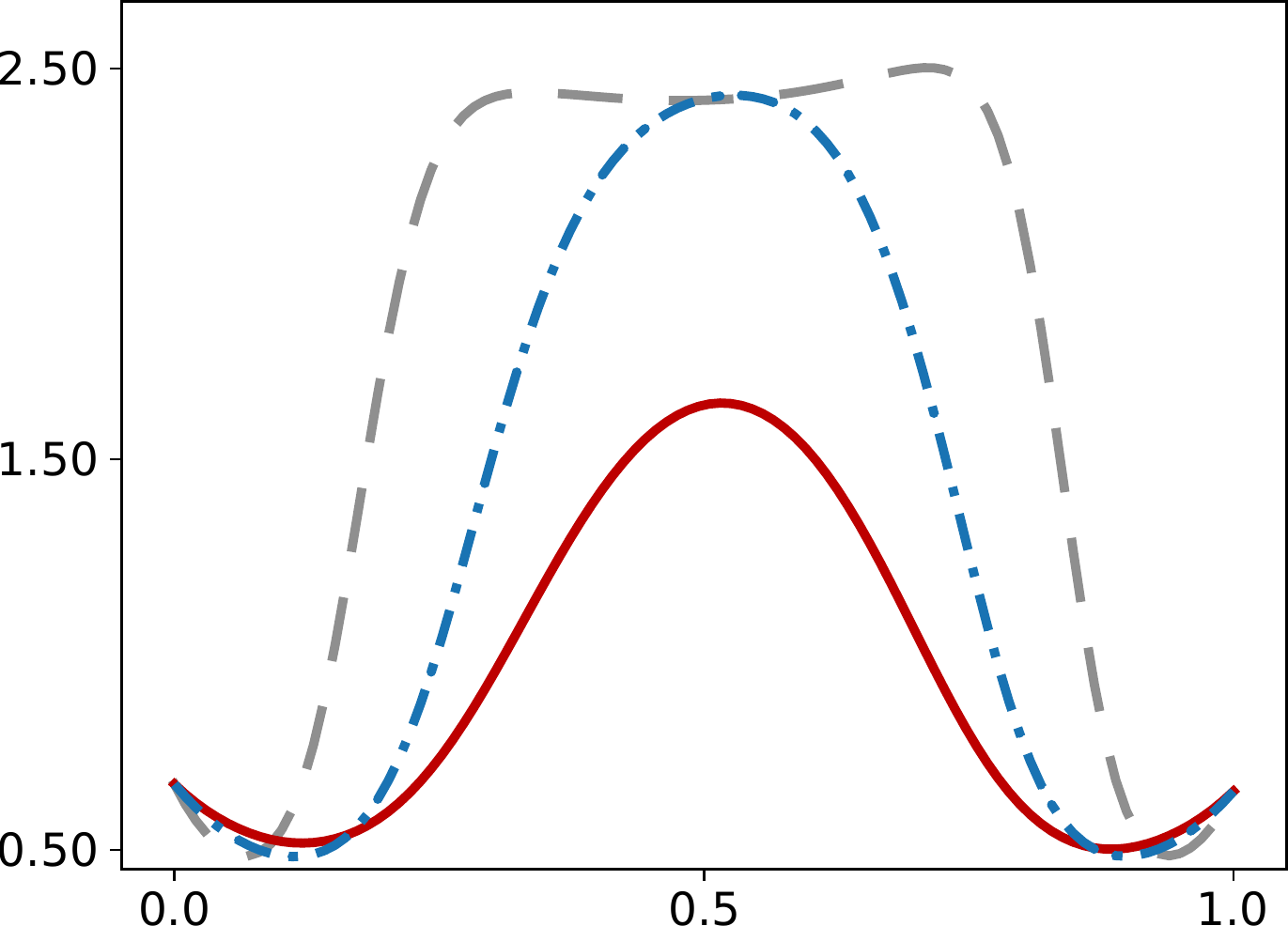} &  &
        \includegraphics[width=.41\linewidth]{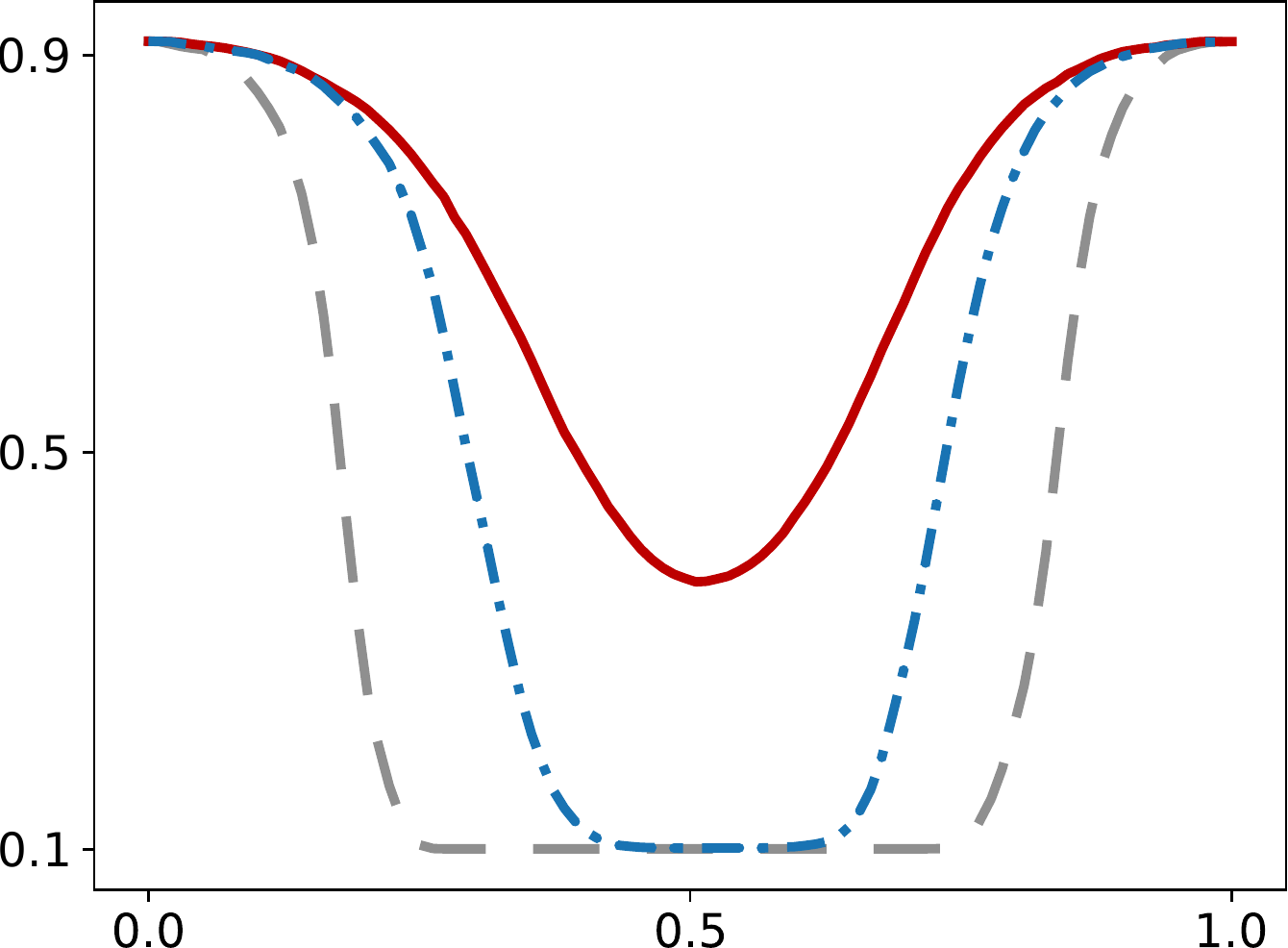}
    \end{tabular}
    \put(-265,144){\small Naive}
    \put(-265,134){\small WM}
    \put(-265,124){\small $\cost_{L2}$}
    \put(-332,-170){$\lambda$}
    \put(-102,-170){$\lambda$}
    \put(-455,75){\rotatebox{90}{loss}}
    \put(-220,68){\rotatebox{90}{accuracy}}
    \put(-455,-90){\rotatebox{90}{loss}}
    \put(-220,-98){\rotatebox{90}{accuracy}}
    \caption{Linear mode connectivity using VGG11 and VGG19 network trained Cifar10 dataset. The left panel shows the loss over the linear path, while the right panel presents the accuracy.}
    \label{fig:lmcvgg11}
\end{figure*}

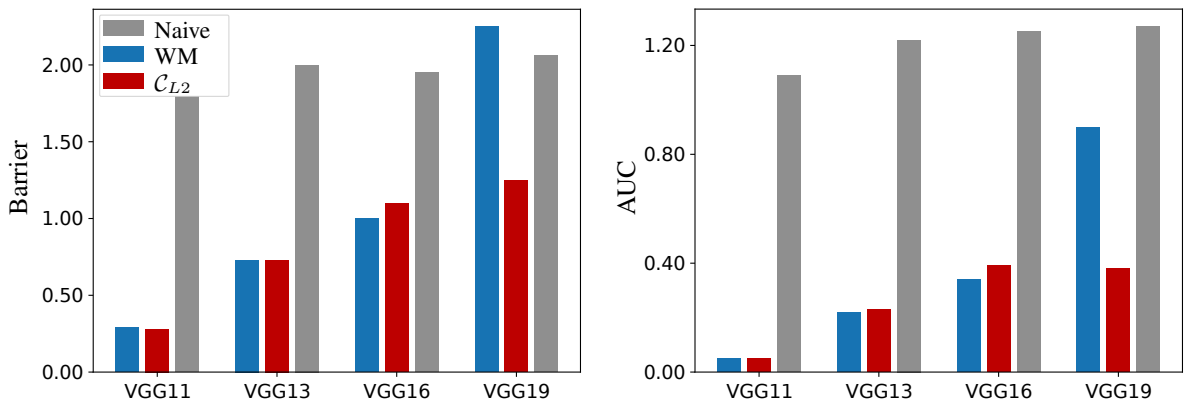
\begin{figure*}[t!]
    \centering
    \begin{tabular}{ccc}
        \includegraphics[width=.41\linewidth]{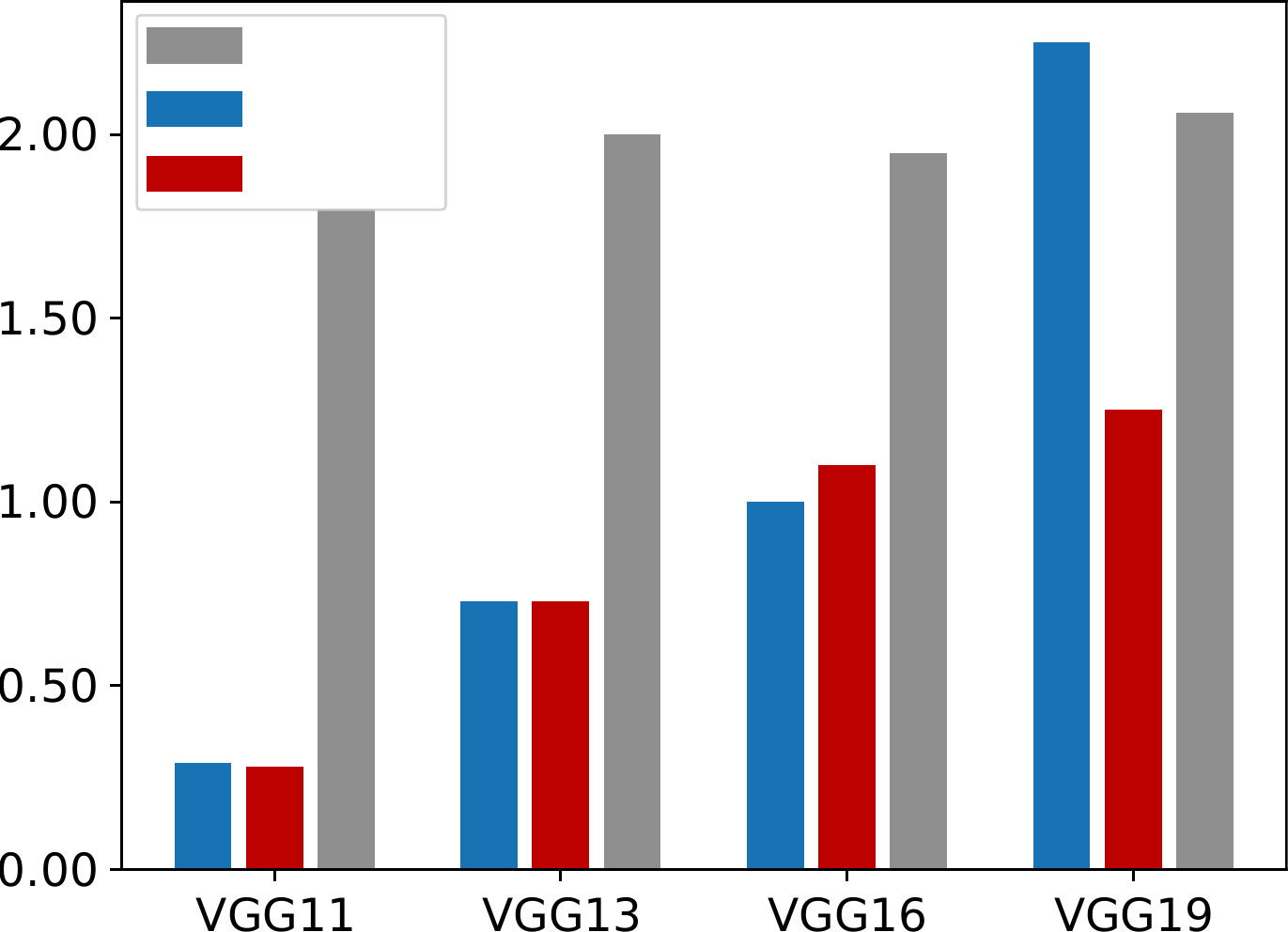} &  &
        \includegraphics[width=.41\linewidth]{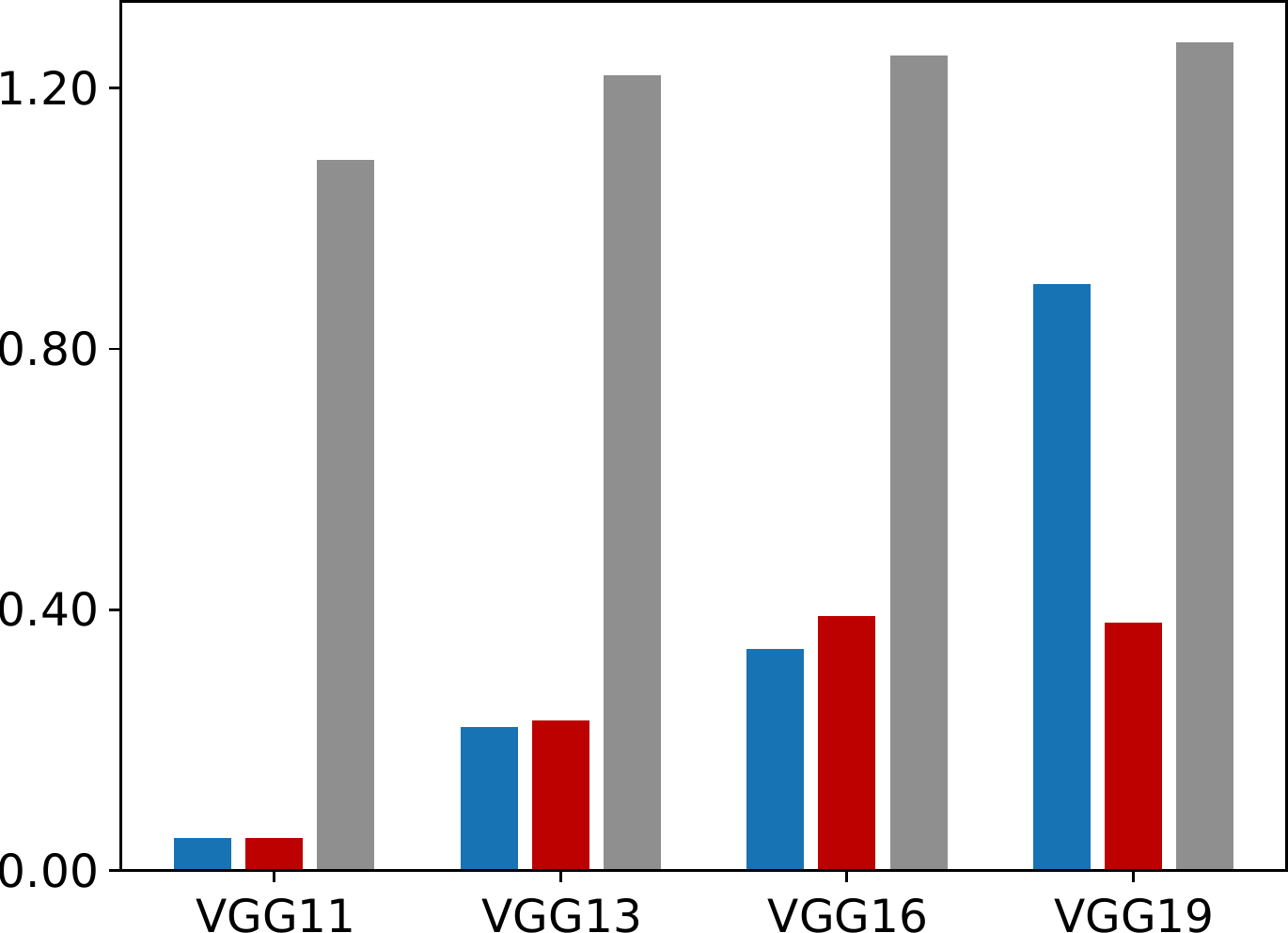}
    \end{tabular}
    \put(-395,67){\small Naive}
    \put(-395,57){\small WM}
    \put(-395,47){\small $\cost_{L2}$}
    \put(-450,0){\rotatebox{90}{Barrier}}
    \put(-220,0){\rotatebox{90}{AUC}}
    \caption{Average Barrier (left panel) and AUC (right panel) of VGG with different depths. }
    \label{fig:lmcgraph}
\end{figure*}

Our method is not limited to NNs for linear mode connectivity as in the previous experiment. Examples of LMC using two VGG11 trained over Cifar10 dataset are shown in \cref{fig:lmcvgg11}. The naive path is presented with dashed gray lines while the cost and accuracy after re-basin with our Sinkhorn network with L2 loss is shown in solid red lines, and Weights Matching (WM)\cite{ainsworth2022git} in dashed blue. An example with VGG19 is also presented in the figure to exemplify some complex cases where LMC is not achieved. Although both methods struggle, our data-free approach can generally find better re-basins. The mean Barrier and AUC of VGGs with different depths are given in \cref{fig:lmcgraph} for our proposal with L2 cost and Weight Matching (WM)\cite{ainsworth2022git}. The code for reproducing this experiment is also provided \href{https://worksheets.codalab.org/bundles/0x5e77ad56e6c24487bbf832fac4d7ba9c}{here}. An example of the execution line using Mnist and NN described in the manuscript is given in this \href{https://worksheets.codalab.org/bundles/0xbed17a1c185e4631a5a1375f90705525}{link}.

\section{Incremental learning algorithm}

\begin{algorithm*}[t!]
    \caption{Re-basin incremental learning}\label{alg:cap}
    \begin{algorithmic}
        \Require A set of episodes $\sT=\{\sT_1,...,\sT_n\}$, and a model $f_{\theta_0}$ trained on $\sT_0$
        \Ensure A model $\theta_n$
        \For{$e=1$ \textbf{to} $n$}
        \State $\mathcal{P}^{0}\gets\{P_j|P_j=\mI\}, 1\leq j\leq h$ \Comment Initialize permutation matrices to identity. $h$ is the number of hidden layers.
        \State $\delta^{0}\gets \text{U}(0,10^{-6})^d$ \Comment Initialize values of vector $\delta^{0}$ following $\text{U}(0,10^{-6})$. $d$ is the number of parameters.
        \For {$i=0$ \textbf{to} $it$}
        \State $(x, y)\gets \text{U}(\sT_e)$ \Comment Sample a mini-batch $(x, y)$ from $\sT_e$.
        \State $\displaystyle\theta^i\gets \frac{\theta_{e-1}+\pi_{\mathcal{P}^i}(\theta_{e-1})}{2}+\delta^{i}$ \Comment Compute the model in the middle of the re-basin line, plus the residual.
        \State $\cost(\delta^{i},\mathcal{P}^{i};\theta^i)=\loss(y,f(x;\theta^i))+\beta||\delta^{i}||^2$ \Comment $\loss$ depends on the task, i.e., cross entropy for classification
        \State $\displaystyle \mathcal{P}^{i+1}\gets\mathcal{P}^{i}-\eta \frac{\partial\cost(\delta^{i},\mathcal{P}^{i};\theta^i)}{\partial \mathcal{P}^{i}}$ \Comment Backpropagation and gradient descent for permutation matrices $\mathcal{P}$
        \State $\displaystyle \delta^{i+1}\gets\delta^{i}-\gamma \frac{\partial\cost(\delta^{i},\mathcal{P}^{i};\theta^i)}{\partial \delta^{i}}$ \Comment Backpropagation and gradient descent for residual $\delta$
        \EndFor
        \State $\theta_e\gets(1-\alpha)\theta_{e-1}+\alpha\theta_{e-1}+\delta^{it}$ \Comment Fuse models for task $\sT_{e-1}$ and $\sT_{e}$
        \EndFor
    \end{algorithmic}
\end{algorithm*}

Incremental learning scenarios followed the standard procedure in the literature, i.e., a benchmark is divided into several episodes, and the new knowledge is incrementally added to the model at episode $i$, $\theta_i$. In our experiments, we employed 20 episodes. Regarding the benchmarks, we used the classical Rotated Mnist, which consists of rotated versions of the Mnist dataset starting with 0 degrees until 180. At each episode, a clockwise rotation of 9.47 degrees was applied to every image in the previous episode. Note that, independently of the rotation applied, the classes remain the same. To address this challenge, we used a feedforward neural network with 1 hidden layer and 256 neurons within the layer. ReLu activation was used for the hidden layer. The input layer size was $784=28\times 28$. The number of neurons in the last layer was 10, corresponding with the number of classes. The algorithm for performing the Re-basin incremental learning is outlined in \cref{alg:cap}. Note that the algorithm is defined using Stochastic Gradient Descent (SGD) for simplicity, but in practice, there is no constraint in which an optimizer can be used. The source code for the re-basin incremental learning is provided \href{https://worksheets.codalab.org/bundles/0x8366fda8885846d5b9ed86da36a63d01}{here}. In our experiments, we used Adam with an initial learning rate  $\eta=0.001$ for the first task and $\eta=0.1$ for the continual learning. As for the residual model, we used SGD with learning rate $\gamma=0.05$ and weight decay $\beta=0.1$. A total of 5 epochs were used to incorporate the new knowledge at each episode using a mini-batch size of 500. As the manuscript relates, the fusion hyper-parameter $\alpha=0.8$ for all our experiments.

As for the second benchmark, we used the Split Cifar100. For such a dataset, we partitioned the Cifar100 dataset into 20 smaller datasets with 5 classes each. Every episode had labeled images corresponding to the 5 categories at hand. The architecture was the multi-head ResNet18 from \cite{mirzadeh2021linear}. Similarly to the previous benchmark, we used Adam for the re-basin and SGD for learning the residual, this time using a mini-batch size of 10 and 20 training epochs for the initial model. The rest of the parameters remained the same as in the previous benchmark, except for the residual vector training, which needed learning rate $\gamma=0.5$ and weight decay $\beta=10^{-4}$.

The accuracy at episode $e=20$ is computed as the average accuracy of model $\theta_e$ over every test dataset from current and previous tasks. On the other hand, forgetting measurement seeks to measure the ability to retain knowledge by computing the highest difference in accuracy between the current model and the previous ones for every task.

\end{document}